\documentclass[10pt,twocolumn,letterpaper]{article}

\usepackage[pagenumbers]{cvpr} 
\usepackage[dvipsnames]{xcolor}
\usepackage{amssymb}
\usepackage{amsmath}
\usepackage{pifont}
\usepackage{multirow}
\usepackage{tabularx}
\usepackage{makecell}
\usepackage{colortbl}
\usepackage{bm}
\usepackage{gensymb}
\usepackage[outline]{contour}
\usepackage{floatrow}
\usepackage[rightcaption]{sidecap}
\usepackage{bbm}
\usepackage{float}
\usepackage[accsupp]{axessibility}  

\usepackage[inkscapelatex=false]{svg}
\definecolor{turquoise}{cmyk}{0.65,0,0.1,0.3}
\definecolor{purple}{rgb}{0.65,0,0.65}
\definecolor{dark_green}{rgb}{0, 0.5, 0}
\definecolor{orange}{rgb}{0.8, 0.6, 0.2}
\definecolor{red}{rgb}{0.8, 0.2, 0.2}
\definecolor{darkred}{rgb}{0.6, 0.1, 0.05}
\definecolor{blueish}{rgb}{0.0, 0.3, .6}
\definecolor{light_gray}{rgb}{0.7, 0.7, .7}
\definecolor{pink}{rgb}{1, 0, 1}
\definecolor{greyblue}{rgb}{0.25, 0.25, 1}
\definecolor{LightRed}{rgb}{0.99,0.89,0.89}

\newcommand{\project}{CrossOver}

\colorlet{colorFst}{Green!25}       
\colorlet{colorSnd}{SpringGreen!45} 
\colorlet{colorTrd}{Yellow!30}      
\colorlet{colorLow}{darkgray!30}
\newcommand{\fs}{\cellcolor{colorFst}\bf}   
\newcommand{\nd}{\cellcolor{colorSnd}}      

\definecolor{teaser_gray}{rgb}{0.88, 0.89, 0.90} 
\definecolor{teaser_green}{rgb}{0.77, 0.88, 0.70} 
\definecolor{teaser_blue}{rgb}{0.70, 0.78, 0.90} 
\definecolor{teaser_yellow}{rgb}{1.0, 0.90, 0.60}

\definecolor{cvprblue}{rgb}{0.21,0.49,0.74}
\definecolor{spidergreen}{RGB}{10,156,10}
\usepackage[
    pagebackref,
    breaklinks,
    colorlinks,
    citecolor=cvprblue
    ]{hyperref}
\usepackage[capitalize]{cleveref}
\title{\project{}: 3D Scene Cross-Modal Alignment}

\author{Sayan Deb Sarkar\textsuperscript{1} \quad
Ondrej Miksik\textsuperscript{2} \quad
Marc Pollefeys\textsuperscript{2, 3} \quad
Daniel Barath\textsuperscript{3, 4} \quad
Iro Armeni\textsuperscript{1} 
\quad
\vspace{5px}
\\
{ \textsuperscript{1}{Stanford University} \quad \textsuperscript{2}{Microsoft Spatial AI Lab}} \quad
\textsuperscript{3}{ETH Zurich}  \quad
\textsuperscript{4}{HUN-REN SZTAKI} \\
 \href{https://sayands.github.io/crossover}{\textcolor{pink}{\texttt{sayands.github.io/crossover}}}
 \\ 
}

\begin{document}
\maketitle
\setlength{\abovedisplayskip}{5pt}
\setlength{\belowdisplayskip}{5pt}
\vspace{-10pt}
\begin{abstract}
Multi-modal 3D object understanding has gained significant attention, yet current approaches often assume complete data availability and rigid alignment across all modalities. We present \project{}, a novel framework for cross-modal 3D \textit{scene} understanding via flexible, scene-level modality alignment. Unlike traditional methods that require aligned modality data for every object instance, \project{} learns a unified, modality-agnostic embedding space for scenes by aligning modalities -- RGB images, point clouds, CAD models, floorplans, and text descriptions -- with relaxed constraints and without explicit object semantics. Leveraging dimensionality-specific encoders, a multi-stage training pipeline, and emergent cross-modal behaviors, \project{} supports robust scene retrieval and object localization, even with missing modalities. Evaluations on ScanNet and 3RScan datasets show its superior performance across diverse metrics, highlighting \project{}’s adaptability for real-world applications in 3D scene understanding. 

\end{abstract}
\vspace{-1.0em}  
\section{Introduction}
\label{sec:intro}   

In recent years, the need to align and transfer information across modalities has grown substantially, especially for tasks involving complex 3D environments. Such a capability enables knowledge and experience transfer across modalities. For example, knowing the layout of kitchens in computer-aided design (CAD) format will provide guidance on how to build a new kitchen, such that it follows the layout of the most similar CAD floorplan.

\begin{figure}
    \includegraphics[width=\linewidth]{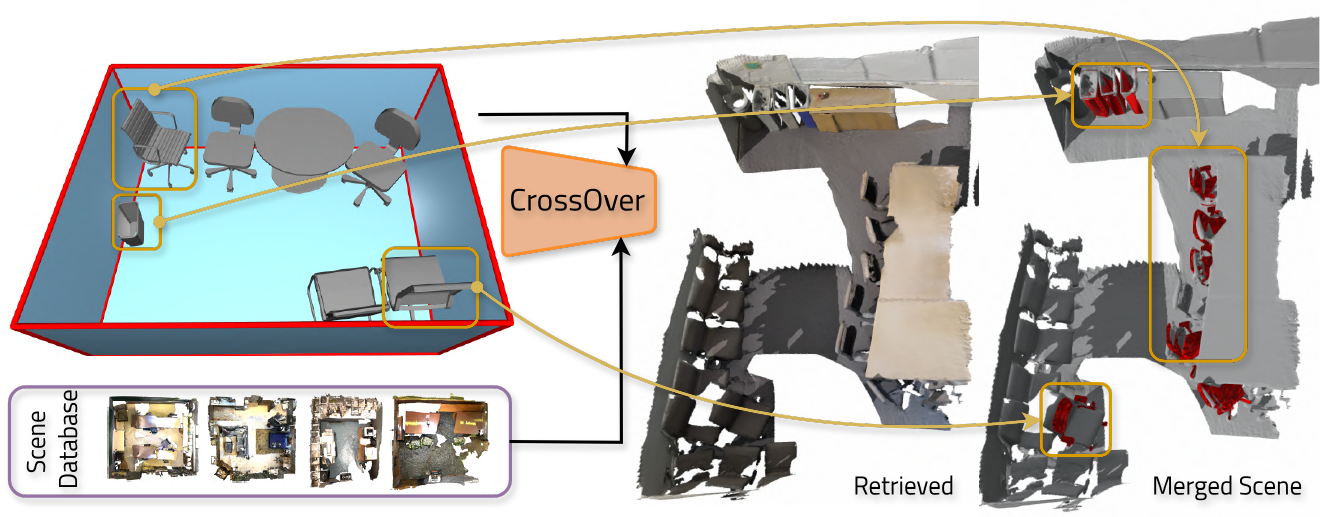}
    \caption{\textbf{\textit{\project{}} is a cross-modal alignment method for 3D scenes that learns a unified, modality-agnostic embedding space, enabling a range of tasks.} For example, given the 3D CAD model of a query scene and a database of reconstructed point clouds, CrossOver can retrieve the closest matching point cloud and, if object instances are known, it can identify the individual locations of furniture CAD models with matched instances in the retrieved point cloud, using brute-force alignment. This capability has direct applications in virtual and augmented reality.}
    \label{fig:teaser}
\end{figure} 

Current multi-modal approaches tackle 3D data alignment of individual objects across modalities \cite{zhang2021pointclip,xue2022ulip,xue2023ulip2,pointbind}, \textit{without including and considering scene context}, making them challenging to extend effectively for scene-level understanding. These methods typically assume fully aligned, consistent datasets, where each modality is perfectly corresponding to all others for each object. However, real-world scenarios rarely provide such complete modality pairings. For example, a video of a room and its CAD model might share some spatial alignment but differ in data characteristics and object instances (hereby referred to as \textit{instances}) represented in the data (e.g., some instances could be missing in one modality, which is common between real-world scenes and their CAD models). Also, achieving consistent instance segmentation across modalities is nearly impossible in practice. Thus, these approaches struggle when certain modalities are missing or incomplete, limiting their flexibility in practical applications \cite{baltruvsaitis2018multimodal}.

We address the inherent limitations of strict object-level modality alignment by introducing a \textit{flexible scene-level} modality alignment approach that operates without prior information during inference (\eg, semantic instance segmentation), unlike the current methods \cite{sarkar2023sgaligner,xie2024sg}. Our method, namely \textbf{\textit{\project{}}} (Fig. ~\ref{fig:teaser}), enables the learning of cross-modal behaviors and relationships, such as identifying similar objects or scenes across different modalities, like the virtual CAD scene based on a video of a real room. This capability extends beyond instance-level matching towards a \textit{unified, modality-agnostic understanding} that supports seamless cross-modal interactions at the scene level.

\textit{\project{}} focuses on aligning five key scene modalities---RGB images, real-world point clouds, CAD models, floorplan images, and text descriptions, in the \textit{feature} space---going beyond the RGB-PC-Text triplets of prior work. Importantly, it is designed with the assumption that not all modalities are available for every data point. By employing a flexible training strategy, we allow \project{} to leverage any available modality during training, without requiring fully aligned data across all modalities. This approach enables our encoders to learn emergent modality alignments, supporting cross-modal traversals even in cases with missing data. Our work is grounded in three key contributions:
\begin{itemize}
\item \textbf{Dimensionality-Specific Encoders:} We introduce 1D, 2D, and 3D encoders tailored to each modality's dimensionality, removing the need for explicit 3D scene graphs or semantic labels during inference. This optimizes feature extraction for each modality and avoids reliance on consistent semantics, which is often hard to obtain.
\item \textbf{Three-Stage Training Pipeline:} Our pipeline progressively builds a modality-agnostic embedding space. First, object-level embeddings capture fine-grained modality relationships. Next, scene-level training develops unified scene representations without requiring all object pairs to align. Finally, dimensionality-specific encoders create semantic-free cross-modal embeddings.
\item \textbf{Emergent Cross-Modal Behavior:} \project{} learns emergent modality behavior, despite not being explicitly trained on all pairwise modalities. It recognizes, \eg, that \textit{Scene$_i$} in the image modality corresponds to \textit{Scene$_i$} in the floorplan modality or its point cloud to the text one, without these modality pairs being present in training.
\end{itemize}

This unified, modality-agnostic embedding space enables diverse tasks such as object localization and cross-modal scene retrieval, offering a flexible, scalable solution for real-world data that may lack complete pairings.
\section{Related Work}
\label{sec:related_work}

\textbf{Multi-modal Representation Learning} aims to bridge data modalities by learning shared embeddings for cross-modal understanding and retrieval.
A seminal work in this area is CLIP~\cite{Radford2021LearningTV}, which popularized the contrastive training objective to learn a joint image-text embedding space. This framework has been extended to various tasks, such as video retrieval~\cite{Luo2021CLIP4Clip}, unified vision-language modeling~\cite{Luo2020UniVL}, and cross-modal alignment~\cite{gao2021clip,Ma2022XCLIP}. In the 3D domain, PointCLIP~\cite{zhang2021pointclip} applied CLIP to point clouds by projecting them into multi-view depth maps, leveraging pretrained 2D knowledge. Subsequent research has focused on multi-modality alignment, \eg ImageBind~\cite{girdhar2023imagebind} aligns six modalities in the 2D domain and shows the power of such representation for generative tasks. In 3D, ULIP~\cite{xue2022ulip} and its successor ULIP-2~\cite{xue2023ulip2} aim to learn unified representations among images, texts, and point clouds. Point-Bind~\cite{pointbind} extends ImageBind~\cite{girdhar2023imagebind} to 3D by aligning specific pairs of modalities using an InfoNCE loss~\cite{oord2018representation}. While these methods effectively capture object-level data, they struggle to differentiate similar instances within a scene, primarily focusing on isolated objects rather than complex scenes. Experiments in Section~\ref{sec:experiment} demonstrate this limitation.

\begin{figure*}[ht!]
    \centering
    \includegraphics[trim=0 0 0 0,clip,width=0.95\linewidth]{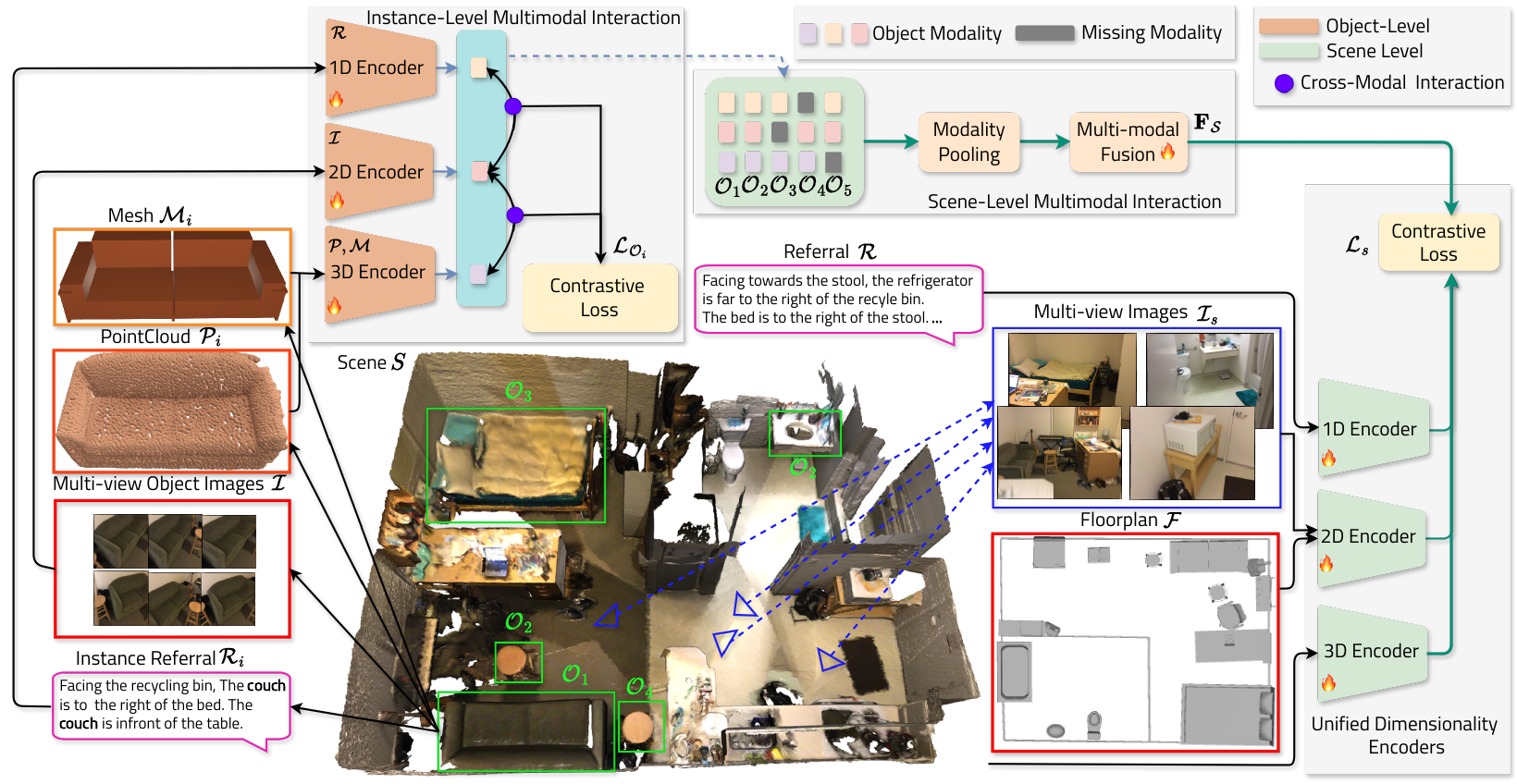}
    \caption{\textbf{Overview of \project{}.} Given a scene $\mathcal{S}$ and its instances $\mathcal{O}_i$ represented across different modalities $\mathcal{I}, \mathcal{P}, \mathcal{M}, \mathcal{R}, \mathcal{F}$, the goal is to align all modalities within a shared embedding space. The \textit{Instance-Level Multimodal Interaction} module captures modality interactions at the instance level within the context of a scene. This is further enhanced by the \textit{Scene-Level Multimodal Interaction} module, which jointly processes all instances to represent the scene with a single feature vector $\mathcal{F_S}$. The \textit{Unified Dimensionality Encoders} eliminate dependency on precise semantic instance information by learning to process each scene modality independently while interacting with $\mathcal{F_S}$.}
    \label{fig:architecture}
    \vspace{-1mm}
\end{figure*}

A common limitation of these approaches is the assumption of \textit{perfect modality alignments} and \textit{complete data} for each instance, often relying on datasets like ShapeNet55~\cite{Chang2015ShapeNetAI}. 
This assumption is impractical for real-world scenarios where data is often incomplete or not well-matched due to occlusions, dynamic changes, sensor limitations, or capture errors, such as in construction sites or robot navigation.
Our work, \project{}, addresses these challenges using real-world datasets consisting of incomplete point clouds and noisy images captured using affordable sensors. Unlike prior methods, we do not require perfect modality alignments or complete data (e.g., point clouds).

\noindent \textbf{3D Scene Understanding} has driven extensive work on text-to-image and point cloud based instance localization and alignment within large maps~\cite{Arandjelovi2015NetVLADCA,camnet2019,text2pos2022}. Techniques like NetVLAD~\cite{Arandjelovi2015NetVLADCA} and CamNet~\cite{camnet2019} enable place recognition and image-based localization by extracting global image descriptors. Recent work has leveraged 3D scene graphs for enhanced scene understanding~\cite{armeni20193d,rosinol20203d,kim20193}, with methods like SGAligner~\cite{sarkar2023sgaligner} and SG-PGM~\cite{xie2024sg} facilitating scene alignment through 3D scene graph matching. For dynamic instance matching across long-term sparse environments, LivingScenes~\cite{zhu2023living} parses an evolving 3D environment with an object-centric formulation. For cross-modal retrieval, approaches like ScanRefer~\cite{chen2020scanrefer} and ReferIt3D~\cite{achlioptas2020referit_3d} localize objects in 3D scenes via natural language but rely on detailed annotations and fixed modality pairs. Methods like 3DSSG~\cite{3DSSG2020} and ``Where Am I"~\cite{Chen2024WhereAI} extend scene retrieval across images and natural language using 3D scene graphs, yet they depend heavily on semantic annotations. SceneGraphLoc~\cite{miao2024scenegraphloc} performs image-to-scene-graph matching, using semantic information. Our approach diverges from these by removing the need for semantics or explicit scene graphs, instead leveraging dimensionality-specific encoders and modality-agnostic embeddings for scene understanding without prior semantic knowledge. 

\noindent \textbf{Handling Missing Modalities} and noisy data is a key challenge in multi-modal learning~\cite{baltruvsaitis2018multimodal}. Traditional approaches often assume full data availability, limiting their real-world applicability. Some methods address missing data through modality imputation or robust models~\cite{tsai2018learning,wu2024comprehensive}. Baltrusaitis \etal~\cite{baltruvsaitis2018multimodal} highlight that many methods lack flexibility for incomplete or noisy data. Our framework tackles this by allowing independent mapping of each modality into a shared embedding space, enabling flexible cross-modal interactions in unstructured environments with sparse or unaligned data. Furthermore, emergent behavior in multi-modal models, such as generalizing and inferring relationships beyond training data~\cite{Radford2021LearningTV,girdhar2023imagebind}, are promoted by structuring training around image embeddings as a common representation. By mapping other modalities into this shared space, \project{} fosters organic cross-modal relationships, enabling unified understanding across diverse data types. 
\section{Method}
\label{sec:method}
Given a 3D scene $\mathcal{S}$ represented by various modalities, denoted as $\mathcal{Q} = \{ \mathcal{I}, \mathcal{P}, \mathcal{M}, \mathcal{R}, \mathcal{F} \}$, our objective is to develop a unified, modality-agnostic representation that maps independent modalities capturing the same 3D scene to a common point in the embedding space. Here, $\mathcal{I}$ is a set of RGB images, $\mathcal{P}$ is a real-world reconstruction as a point cloud, $\mathcal{M}$ is a digital mesh representation from computer aided design (CAD), $\mathcal{R}$ is textual data describing $\mathcal{S}$ within its surroundings, and $\mathcal{F}$ is a rasterized floorplan.

Our proposed framework facilitates robust interactions across different modalities at both the comprising instances and scene levels, enhancing the multi-modal (\eg, pointcloud $\mathcal{P}$ and floorplan $\mathcal{F}$) and same modal (e.g., textual data $\mathcal{R}$) understanding of 3D environments. We structure the development of the embedding space progressively, beginning with instance-level multi-modal interactions and culminating in scene-level multi-modal interactions without requiring prior knowledge, such as semantic information about constituent instances. An overview of \project{} is shown in Fig. \ref{fig:architecture}. To demonstrate the capabilities of this unified, modality-agnostic embedding space, we evaluate:
\begin{enumerate}
    \item{\textbf{Cross-modal instance retrieval}}: Given an observed modality $\mathcal{Q}_{j}$ of a query instance $\mathcal{O}_i$ in a scene $\mathcal{S}$ (\eg, mesh $\mathcal{M}$ or pointcloud $\mathcal{P}$), we aim to retrieve any other modality $\mathcal{Q}_{k}$ representing $\mathcal{O}_i$ within $\mathcal{S}$.
    \item{\textbf{Cross-modal scene retrieval}}: Given a scene $\mathcal{S}_i$ represented by modality $\mathcal{Q}_j$ (\eg, image $\mathcal{I}$ or floorplan $\mathcal{F}$), we aim to retrieve another modality $\mathcal{Q}_{k}$ representing $\mathcal{S}_i$.
\end{enumerate}

\subsection{Instance-Level Multi-Modal Interactions}
\label{sec:object_encoder_training}

First, we describe the pipeline used for learning a multi-modal embedding space for independent instances.
This will provide a basis for the scene-level embeddings. 
We process each of the 1D ($\mathcal{R}$), 2D ($\mathcal{I}$), and 3D ($\mathcal{P}$ and $\mathcal{M}$) instance modalities with corresponding encoders\footnote{The $\mathcal{F}$ modality is not used when learning an instance-level embedding since there is no notion of a floorplan in this scenario.}:

\noindent \textbf{1D Encoder.} 
An instance $\mathcal{O}_i$ can be represented by its textual context in a scene $\mathcal{S}$, using descriptions like \textit{``The chair is in front of the lamp"} and \textit{``The chair is left of the table"}. We term these descriptions as \textit{object referrals}~\cite{jia2024sceneverse} and encode each referral as $f_{ij}^\mathcal{R}$ using the pre-trained text encoder BLIP~\cite{Li2022BLIPBL}, where $i$ is the instance of interest (\eg, chair) and $j$ represents another instance in the scene (\eg, lamp, table, or another chair). Practically, we collect $k$ object referrals per instance, resulting in $F_i^\mathcal{R} = \{ f_{i1}^{\mathcal{R}}, \ldots, f_{ik}^{\mathcal{R}} \}$. To create a single feature vector $f_{i}^{\mathcal{R}}$ representing the instance’s context, we apply average pooling over $F_i^\mathcal{R}$ .

\noindent \textbf{2D Encoder.} Given a collection $I_\mathcal{S}$ of images capturing a scene $\mathcal{S}$, we integrate multi-view and per-view multi-level visual embeddings for each $\mathcal{O}_i$ to encode $f_i^\mathcal{I}$. Inspired by~\cite{miao2024scenegraphloc}, for each $\mathcal{O}_i$, we select the top $K_{view}$ defined by largest visibility of  $\mathcal{O}_i$ among $I_\mathcal{S}$ and calculate multi-level bounding boxes around $\mathcal{O}_i$ $\{b_{v,l} \; | \; l \in [0, L)\}$ within each view $v$. A pre-trained DinoV2~\cite{oquab2023dinov2, darcet2023vitneedreg} encoder processes the image crops defined by $b_{v,l}$ to give us the \texttt{[CLS]} tokens per crop~\cite{yang2024denoising}. Subsequent average pooling operations aggregate these tokens into a singular feature vector $f_i^\mathcal{I}$. In contrast to~\cite{miao2024scenegraphloc}, we do not assume available camera poses.

\noindent \textbf{3D Encoder.} Given instance $\mathcal{O}_i$ and its corresponding real-world \textit{point cloud} $\mathcal{P}_i$ and \textit{shape mesh} $\mathcal{M}_i$, we extract instance features $\Bar{f_i^\mathcal{P}}$ and $\Bar{f_i^\mathcal{M}}$ using a pretrained I2PMAE~\cite{Zhang2022Learning3R} point cloud encoder. Importantly, we do not utilize the semantic class~\cite{jia2024sceneverse, 3dvista} of $\mathcal{O}_i$ in these operations. We concatenate the 3D location of $\mathcal{P}_i$ and $\mathcal{M}_i$ to $\Bar{f_i^\mathcal{P}}$ and $\Bar{f_i^\mathcal{M}}$, respectively, to form the instance tokens $\hat{f_i^\mathcal{P}}$ and $\hat{f_i^\mathcal{M}}$. To introduce partial scene-level reasoning, we incorporate interactions between instances by integrating the instance tokens and encoding the pairwise spatial relationships of an instance with all others in $\mathcal{S}$ within a transformer network. Similar to~\cite{jia2024sceneverse}, we employ spatial-attention-based transformers, following~\cite{3dvista, chenlanguage2022}, to generate $f_i^\mathcal{P}$ and $f_i^\mathcal{M}$. Details about the 3D location and spatial relationships are in Supp. For the mesh modality $\mathcal{M}$, we sample points on the mesh surface to enable input to a point cloud encoder. We encode neither the 3D location nor the spatial pairwise relation among instances, as we do not assume that the meshes are aligned with the scene geometry.

All pre-trained encoders, which are frozen during training, are followed by trainable projection layers. During training, after encoding each modality, we apply a contrastive loss to enforce alignment of modality features within a joint embedding space. Unlike prior work that requires full data modality alignment~\cite{pointbind, xue2023ulip2} or semantic scene graph \cite{sarkar2023sgaligner,miao2024scenegraphloc}, \project{} accommodates the practical challenge that not all modalities may always be available by not requiring the presence of all modalities simultaneously. Instead, it aligns all other modality embeddings with image space $\mathcal{I}$. The loss function can be defined as:
\begin{equation}
\vspace{-2pt} 
    \mathcal{L}_{\mathcal{O}_i} = \mathcal{L}_{f_i^I, f_i^\mathcal{P}} + \mathcal{L}_{f_i^I, f_i^\mathcal{M}} + \mathcal{L}_{f_i^I, f_i^\mathcal{R}}.
\end{equation}

 During \textit{training}, \project{} requires a base modality for every instance, to align other modalities with its feature space. We choose images $\mathcal{I}$ as the base modality due to their availability and strong encoder priors, though any supported modality can serve this role. Crucially, \textit{no modality availability assumptions are made during inference}, allowing any query-target modality pair. Our experiments (see Supp.) show that aligning to a single reference modality, rather than using all pairwise combinations as in prior work, improves performance.

\subsection{Scene-Level Multi-Modal Interactions}
\label{sec:scene_encoder_training}
We distill knowledge from instance-level modality encoders to scene-level encoders, allowing us to leverage instance-based insights during training and enabling scene-level retrieval at inference without relying on 3D scene graphs or semantic instance information across modalities.

\noindent \textbf{Multi-modal Scene Fusion.} Given the instance features $f_i^\mathcal{R}$, $f_i^\mathcal{I}$, $f_i^\mathcal{P}$, and $f_i^\mathcal{M}$ for each instance $\mathcal{O}_i$ in scene $\mathcal{S}$, we compute each of the scene level features $f^\mathcal{R}$, $f^\mathcal{I}$, $f^\mathcal{P}$, and $f^\mathcal{M}$ by first applying average pooling per modality to the features of all instances in $\mathcal{S}$. We then perform a weighted fusion of these pooled features to learn a fixed-size multi-modal embedding $\mathbf{F}_\mathcal{S}$:
\begin{equation}
\mathbf{F}_{\mathcal{S}} = \sum_{q \in \mathcal{Q}} \left[ \frac{\exp(w_q)}{\sum_{j \in \mathcal{Q} \setminus q} \exp(w_j)} f^q \right],
\end{equation}
where  $j, q \in \mathcal{Q}$, $w_q$ and $w_j$ are modality-wise trainable attention weights. We use an MLP head to project the dimensionality to our final representation space, resulting in an embedding that serves as a unified scene representation, capturing interactions across all modalities. In practice, this representation is flexible, adapting to data availability and specifically to any missing modalities.

\subsection{Unified Dimensionality Encoders}
\label{sec:unified_encoder_training}

The above scene-level encoder provides a unified, modality-agnostic embedding space; however, it requires semantic instance information consistent across modalities during inference, which is challenging to obtain in practice. To eliminate this need, we design a single encoder per modality dimensionality (\ie 1D, 2D, and 3D) that directly processes raw data without needing additional information. Moreover,
our experiments (Supp.) show that the scene-level encoder
needs all modalities at inference to perform reasonably. 

\noindent \textbf{1D Encoder.} Similar to Sec. \ref{sec:object_encoder_training}, we use \textit{object referrals} to describe scene context~\cite{3dvista}. We randomly sample $t=10$ referrals per scene and use a text encoder to form $\mathbf{F}_{1D}$.

\noindent \textbf{2D Encoder.} Here, we consider both RGB and floorplan images. The floorplan $\mathcal{F}$ is represented as a top-view orthographic projection image of the $3D$ layout with geometrically aligned shape meshes for furniture instances. Since a scene can be captured with multiple RGB images $\mathcal{I}_\mathcal{S}$, we employ a naive key-frame selection strategy to sample $N=10$ multi-view images (see Supp.). We process the images using a DinoV2~\cite{oquab2023dinov2} encoder and concatenate the output \texttt{[CLS]} token and aggregated patch embeddings to form $\mathbf{F}_{2D}^i, i \in N$. We pass each $\mathbf{F}_{2D}^i$ via an MLP projection head and apply average pooling to generate $F_{2D}$. In practice, we use the same encoder with shared weights for both RGB images $\mathcal{I}_\mathcal{S}$ and floorplan $\mathcal{F}$; \ie, inputs are not distinguished between RGB and floorplan during training. This is the first use of the floorplan modality in \project{} and there is \textit{no pairwise modality interaction} during training between it and the image modality, unlike other modalities.

\noindent \textbf{3D Encoder.} We utilize a sparse convolutional architecture with a residual network as the encoder, built with the Minkowski Engine~\cite{choy20194d}. Given an input point cloud $P \in \mathbb{R}^{N \times 3}$ containing $N$ points, it is first quantized into $M_0$ voxels represented as $V \in \mathbb{R}^{M_0 \times 3}$. The model then produces a full-resolution output feature map $\mathbf{F}_{3D} \in \mathbb{R}^{M_0 \times D}$.

\par The goal is to align each of the unified dimensionality encoders with the scene-level multi-modal encoder. The loss function for unified training becomes:
\begin{equation}
    \mathcal{L}_s = \alpha \mathcal{L}_{\mathbf{F}_{\mathcal{S}}, \mathbf{F}_{1D}} + \beta \mathcal{L}_{\mathbf{F}_{\mathcal{S}}, \mathbf{F}_{2D}} + \gamma \mathcal{L}_{\mathbf{F}_{\mathcal{S}}, \mathbf{F}_{3D}},
\end{equation}
where, $\alpha$, $\beta$, and $\gamma$ are learnable hyper-parameters.

Thus, our combined loss is as follows:
\begin{equation}
    \mathcal{L} = \mathcal{L}_s + \sum_{\mathcal{O}_i \in \mathcal{S}} \mathcal{L}_{\mathcal{O}_i}
    \label{eq:loss_objective}
\end{equation}

\subsection{Loss Definition} Given $q = G({\mathcal{Q}^m}_{i})$ and $k = H({\mathcal{Q}^n}_i)$, $i \in \mathcal{B}$, two different encoder outputs for modalities $\mathcal{Q}^m$ and $\mathcal{Q}^n$ in minibatch $\mathcal{B}$, we use a contrastive loss similar to \cite{girdhar2023imagebind}:
\begin{equation}
    \vspace{-4pt}
    \mathcal{L}_{q, k} = - \log \frac{exp(q_i^Tk_i / \tau)}{exp(q_i^Tk_i / \tau) +  \sum_{j \neq i}{exp(q_i^Tk_j / \tau)}}.
\end{equation}
Here, $\tau$ is a learnable temperature parameter, to modulate similarity between positive pairs. We consider every example $j \neq i$ in a minibatch $\mathcal{B}$ as a negative example. In practice, we use a symmetric loss for better convergence: $\mathcal{L}_{q, k} + \mathcal{L}_{k, q}$. Although we pair each modality with the most prevalent one (\ie, $\mathcal{I}$) to avoid the need for fully aligned modalities per data point during training, there are cases where not all modality pairs are available for a given data point. To enhance \project{}'s flexibility, we account for these scenarios by masking the corresponding loss term for any unavailable modality pairs. 

\subsection{Inference}
\begin{figure}
    \centering
    \includegraphics[width=\linewidth]{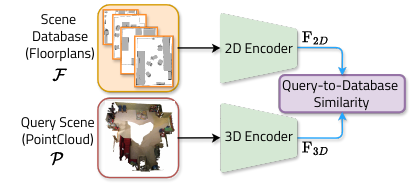}
    \caption{\textbf{Cross-modal Scene Retrieval Inference Pipeline.} Given a query modality ($\mathcal{P}$) that represents a scene, we obtain with the corresponding dimensionality encoder its feature vector ($\mathcal{F}_{3D}$) in the shared cross-modal embedding space. We identify the closest feature vector ($\mathcal{F}_{2D}$) in the target modality ($\mathcal{F}$) and retrieve the corresponding scene from a database of scenes in $\mathcal{F}$.}  
    \label{fig:cross_modal_retrieval_schema}
\end{figure} 
\vspace{-1mm}

After training \project{} with the loss objective defined in Eq. \ref{eq:loss_objective}, we use the embedding feature vectors for retrieval tasks. Given a scene $S$ containing $\mathcal{O} = \{ \mathcal{O}_i \}$ instances each represented by one or more modalities from $\mathcal{Q}$, we use our instance-level multi-modal encoders to perform cross-modal retrieval. Given $\mathcal{O}_i$ in query modality $\mathcal{Q}_j$ and all other instances in target modality $\mathcal{Q}_k$, the goal is to retrieve the $\mathcal{O}_i$ in $\mathcal{Q}_k$. For scene retrieval, we apply a similar approach using our unified dimensionality encoders, except that instead of instances, we retrieve entire scenes. A schematic diagram for one modality pair is shown in Fig.~\ref{fig:cross_modal_retrieval_schema}.

\begin{figure*}[t]
  \begin{subfigure}[b]{0.44\linewidth}
    \centering
    \includegraphics[width=\textwidth]{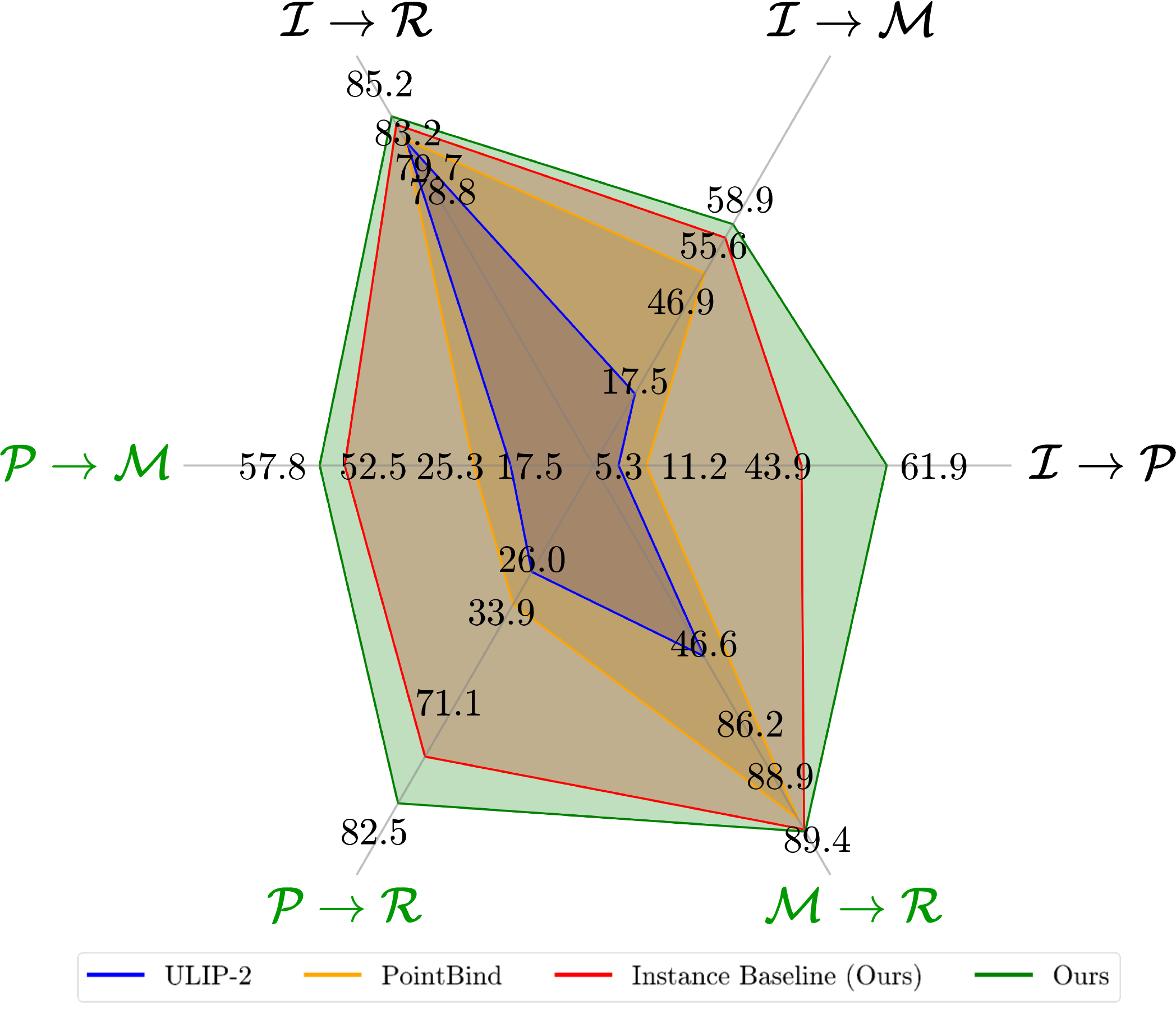}
    \caption{\textbf{Instance Matching Recall on ScanNet}}
    \label{fig:spider_instancematching}
  \end{subfigure}
  \hfill
  \begin{subfigure}[b]{0.5\linewidth}
    \centering
    \resizebox{\linewidth}{!}{
    \begin{tabular}{l|ccc|ccc}
    \toprule
       &  \multicolumn{3}{c}{\textbf{Scannet}~\cite{dai2017scannet}} & \multicolumn{3}{c}{\textbf{3RScan}~\cite{wald2019rio}} \\
     \midrule\arrayrulecolor{black} 
       \textbf{Scene-level Recall} $\uparrow$ & \textbf{R@25}\% & \textbf{R@50}\% & \textbf{R@75}\% & \textbf{R@25}\% & \textbf{R@50}\% & \textbf{R@75}\%  \\
    \midrule
    \multicolumn{7}{l}{\cellcolor[HTML]{EEEEEE}{\textit{$\mathcal{I} \rightarrow \mathcal{P}$}}} \\
    ULIP-2 ~\cite{xue2023ulip2} & 1.28 & 0.64 & 0.24 & 1.91 & 0.40 & 0.28 \\
    PointBind ~\cite{pointbind} & 6.73 & 0.96 & 0.32 & 3.18 & 0.64 & 0.01 \\
    Inst.\ Baseline (Ours) & \nd 88.46 & \nd 37.82 & \nd 1.92 & \nd 93.63 & \nd 35.03 & \nd 3.82 \\
    Ours & \fs 98.08 & \fs 76.92 & \fs 23.40 & \fs 99.36 & \fs 79.62 & \fs 22.93 \\
    \multicolumn{7}{l}{\cellcolor[HTML]{EEEEEE}{\textit{$\mathcal{I} \rightarrow \mathcal{R}$}}} \\
    ULIP-2 ~\cite{xue2023ulip2} & 98.12 & 96.21 & 60.34 & 98.66 & 85.91 & 36.91 \\
    PointBind ~\cite{pointbind} & 98.22 & 95.17 & 62.07 & \fs 100 & 87.25 & 41.61 \\
    Inst.\ Baseline (Ours) & \nd 99.31 & \nd 97.59 & \nd 71.13 & \fs 100	& \nd 92.62	& \nd 55.03 \\
    Ours & \fs 99.66 & \fs 98.28 & \fs 76.29 & \fs 100 & \fs 97.32 & \fs 67.79 \\
    \multicolumn{7}{l}{\cellcolor[HTML]{EEEEEE}{\textit{$\mathcal{P} \rightarrow \mathcal{R}$}}} \\
    ULIP-2 ~\cite{xue2023ulip2} & 37.24 & 16.90 & 8.62 & 16.78 & 6.04 & 1.34 \\
    PointBind ~\cite{pointbind} & 54.83 & 27.93 & 11.72 & 21.48 & 6.04 & 2.01 \\
    Inst.\ Baseline (Ours) & \nd 98.63 & \nd 83.85 & \nd 46.74 & \nd 92.62	& \nd 60.40 & \nd 20.81 \\
    Ours & \fs 99.31 & \fs 96.56 & \fs 70.10 & \fs 100 & \fs 89.26 & \fs 50.34 \\
    \bottomrule
    \end{tabular}
}
    \vspace{2em}
  \caption{\textbf{Scene-Level Matching Recall on ScanNet and 3RScan}}
  \label{fig:table_scene_level_instance_matching}
  \end{subfigure}
  \caption{
  \textbf{Cross-Modal Instance Retrieval on ScanNet and 3RScan.} (a) Even though \project{} does not explicitly train all modality combinations, it achieves emergent behavior within the embedding space. The same applies to our Instance Baseline (Ours). \project{} performs better than our self-baseline since it incorporates more scene context in the fusion of modalities. (b) Our method outperforms all baselines in all datasets, showcasing the robustness of learned cross-modal interactions.
}
  \label{fig:instance_matching}
\end{figure*}
\section{Experiments}
\label{sec:experiment}

\textbf{Datasets.}
We train and evaluate \project{} on ScanNet \cite{dai2017scannet} and 3RScan~\cite{wald2019rio}. We choose ScanNet for providing comprehensive coverage of all modalities, and 3RScan for including more data on temporal scenes. For both, we use the \textit{object referrals} from SceneVerse~\cite{jia2024sceneverse}, which is a million-scale 3D vision-language dataset with $68$K 3D indoor scenes comprising indoor scene understanding datasets and $2.5M$ vision-language pairs. In all evaluations, we use a model trained across all datasets (details in Supp.).

\textbf{ScanNet \cite{dai2017scannet}} is an RGB-D video dataset containing 2.5 million views in more than $1500$ scenes, annotated with 3D camera poses, surface reconstructions, and instance-level semantic segmentation; we obtain images and 3D point clouds. For mesh $\mathcal{M}$ and floorplan $\mathcal{F}$, we use the Scan2CAD~\cite{Avetisyan_2019_CVPR} dataset, which provides annotated keypoint pairs between CAD models from ShapeNet~\cite{Chang2015ShapeNetAI} and their counterpart objects in the scans. \textbf{3RScan \cite{wald2019rio}} benchmarks instance relocalization, featuring $1428$ RGB-D sequences across $478$ indoor scenes, including rescans of the latter after object relocation. It provides annotated 2D and 3D instance segmentation, camera trajectories, and reconstructed scan meshes. We obtain images and point clouds.

\begin{figure*}
    \centering
    \includegraphics[width=0.8\linewidth]{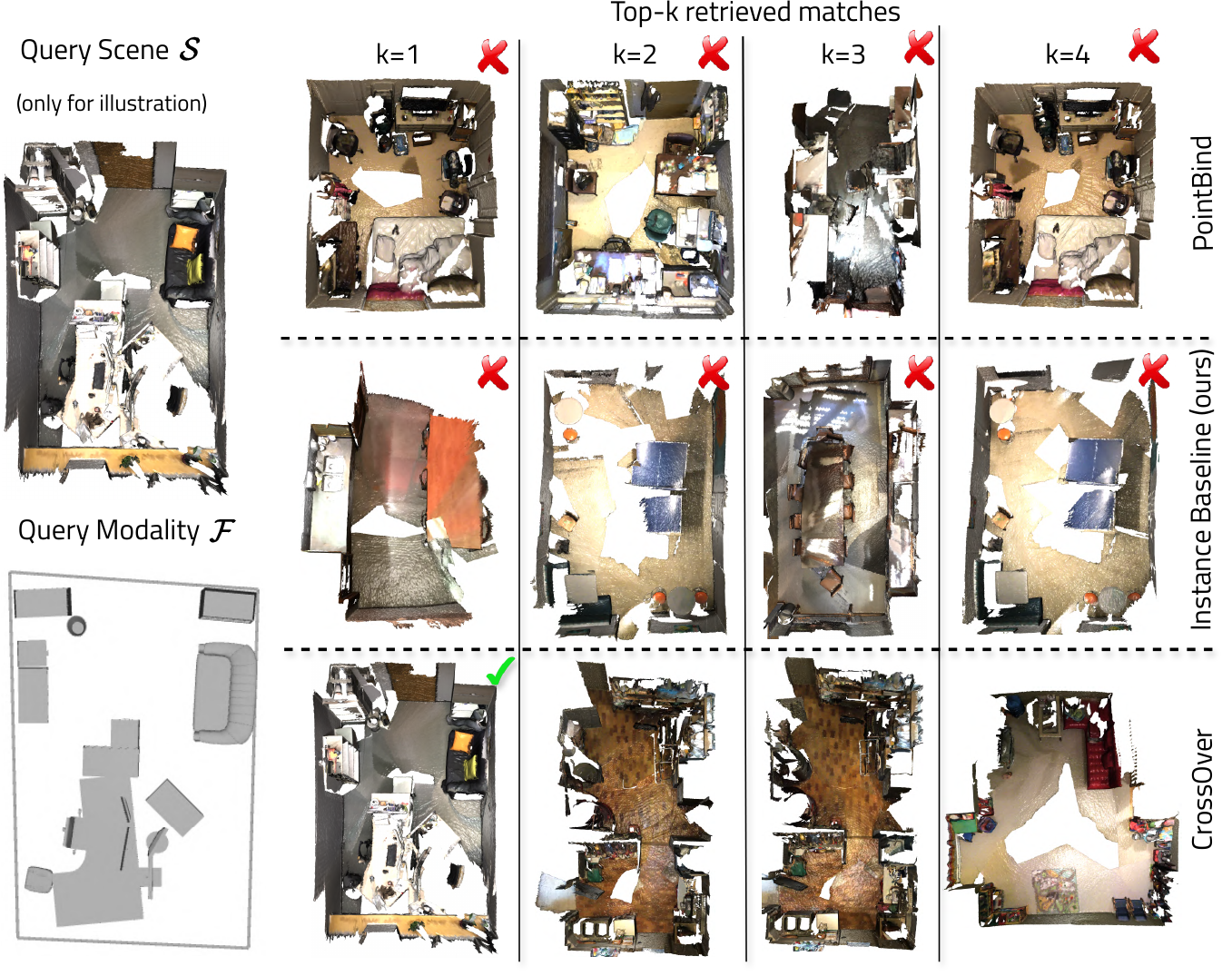}
    \caption{\textbf{Cross-Modal Scene Retrieval Qualitative Results on ScanNet.} Given a scene in query modality $\mathcal{F}$, we aim to retrieve the same scene in target modality $\mathcal{P}$. While PointBind and the Instance Baseline do not retrieve the correct scene within the top-4 matches, \project{} identifies it as the top-1 match. Notably, temporal scenes appear close together in \project{}’s embedding space (\eg, $k=2$, $k=3$), with retrieved scenes featuring similar object layouts to the query scene, such as the red couch in $k=4$.}
    \label{fig:visual_comparison}
    \vspace{-10pt}
\end{figure*} 

\noindent
\textbf{Evaluation Metrics.}
We assess the quality of our representation by quantifying its ability to identify the same instance $\mathcal{O}_i$ or scene $\mathcal{S}_i$ across modalities, $\mathcal{Q}_j$ and $\mathcal{Q}_k$. Extending image feature matching evaluation~\cite{lowe2004, sarlin20superglue}, we compute the \textit{instance matching recall} as the ratio of correctly identified $\mathcal{O}_i$ matches, given a database of instances. Additionally, we evaluate \textit{scene-level (instance) matching recall} at thresholds of $25\%$, $50\%$, and $75\%$, indicating how many objects from a scene in modality $\mathcal{Q}_j$ out of the total objects in the same scene we can match in modality $\mathcal{Q}_k$. This combined measure shows instance matching failure within a scene.

We further evaluate the challenging task of cross-modal scene retrieval within a database. For example, given a query point cloud of a scene, we aim to retrieve its corresponding 2D floorplan. This analysis includes multiple levels: (i) \textit{scene matching recall}, or the model's ability to retrieve the exact scene $\mathcal{S}_i$; (ii) \textit{scene category recall} to test retrieval of a scene from the same category (\eg, retrieving \textit{any} kitchen when given a kitchen query in a multi-category database); (iii) \textit{temporal recall} to evaluate whether the model can recover the same scene captured at a different time, accounting for potential object movement or removal; and (iv) \textit{intra-category recall}, which assesses retrieval of a specific scene within a single-category database (\eg, retrieving a particular kitchen from only kitchen scenes). This last metric uniquely requires a different database.

\subsection{Instance Retrieval}
\noindent \textbf{Cross-Modal Instance Matching.}  Our goal is instance matching within the same scene where multiple instances of the same furniture (\eg, \textit{two identical chairs}) are commonly present. We showcase our results on \textit{ScanNet} and \textit{3RScan} datasets in Fig.~\ref{fig:instance_matching}. We compare \project{} with pretrained multi-modal methods ULIP-2~\cite{xue2023ulip2} and PointBind~\cite{pointbind} and our instance-level multi-modal encoder to highlight the importance of scene-level understanding in a cross-modal embedding space. As shown in Fig.~\ref{fig:spider_instancematching}, our performance on ScanNet is robust across modalities, while baselines exhibit varying results. Current multi-modal methods are large pretrained models with strong text encoders that boost performance for \textit{referral}-based retrieval. While prior work trains on all pairwise modalities, we selectively train only in reference to the image modality ($\mathcal{I}$). Yet, we still achieve robust performance across all modalities, even without direct interactions during training. Emergent interactions are in \textcolor{spidergreen}{green}. Similar trends appear in Fig. \ref{fig:table_scene_level_instance_matching} for scene-level matching.

\begin{table}[t]
  \centering
  \resizebox{0.80\linewidth}{!}{
   \begin{tabular}{l|ccc}
    \toprule
    \multicolumn{4}{c}{\textbf{Scene-level Recall} $\uparrow$} \\  
     \midrule\arrayrulecolor{black}
      \multicolumn{1}{l|}{\textbf{Method}} & \textbf{R@25}\% & \textbf{R@50}\% & \textbf{R@75}\%  \\
    \midrule
    \multicolumn{4}{l}{\cellcolor[HTML]{EEEEEE}{\textit{same-modal ($\mathcal{P} \rightarrow \mathcal{P}$)}}} \\
    MendNet~\cite{duggal2022mending} & 80.68 & 64.77 & 37.50\\
    VN-DGCNN$_{\mathrm{cls}}$~\cite{deng2021vector} & 72.32&53.41 & 29.55 \\
    VN-ONet$_{\mathrm{recon}}$~\cite{deng2021vector} & 86.36 & 71.59 & 44.32\\
    LivingScenes ~\cite{zhu2023living} & \nd 87.50 & \nd 78.41 & \nd 50.00 \\
    Ours & \fs 92.31 & \fs 84.62 & \fs 57.69 \\
    \multicolumn{4}{l}{\cellcolor[HTML]{EEEEEE}{\textit{cross-modal (ours) }}} \\
    $\mathcal{I} \rightarrow \mathcal{P}$ & 89.74 & 73.08 & 42.31 \\
    $\mathcal{I} \rightarrow \mathcal{R}$ & 62.33 & 38.96 & 18.18 \\
    $\mathcal{P} \rightarrow \mathcal{R}$ & 68.83 & 40.26 & 22.08 \\
    \bottomrule
    \end{tabular}
    }\vspace{-5pt}
	\caption{\textbf{Temporal Instance Matching on \emph{3RScan}~\cite{wald2019rio}.} Our method exhibits better performance in the same-modal task compared to baselines, despite not being specifically trained on this. It also performs well on cross-modal tasks. Lower performance when $\mathcal{R}$ is involved is expected, as descriptions are contextualized within the scene's layout and may lose validity if objects rearrange.}
	\label{tab:temporal_instance_matching_3rscan}
\end{table}

\noindent \textbf{Temporal Instance Matching.} Although not part of the learning objective, we evaluate \project{}'s effectiveness on temporal point cloud-based instance retrieval (same-modal) using scans acquired at different time intervals, with scene changes like object displacement and rearrangement. Tab.~\ref{tab:temporal_instance_matching_3rscan} shows a comparison on the \textit{3RScan} dataset, highlighting our method's superior performance. This is a large gain, lying in the strong representational power of our multi-modal embedding space, which allows the encoder to efficiently extract each instance's spatial and geometric features in dynamic scenes. Moreover, our method, while primarily evaluated in the same-modal setting, also demonstrates superior performance in the cross-modal scenario, shown in the second half of Tab.~\ref{tab:temporal_instance_matching_3rscan}, further underlining the importance of scene-level multi-modal alignment to handle temporal variations in indoor scene understanding.

\begin{figure*}
    \centering
    \includegraphics[width=0.95\columnwidth]{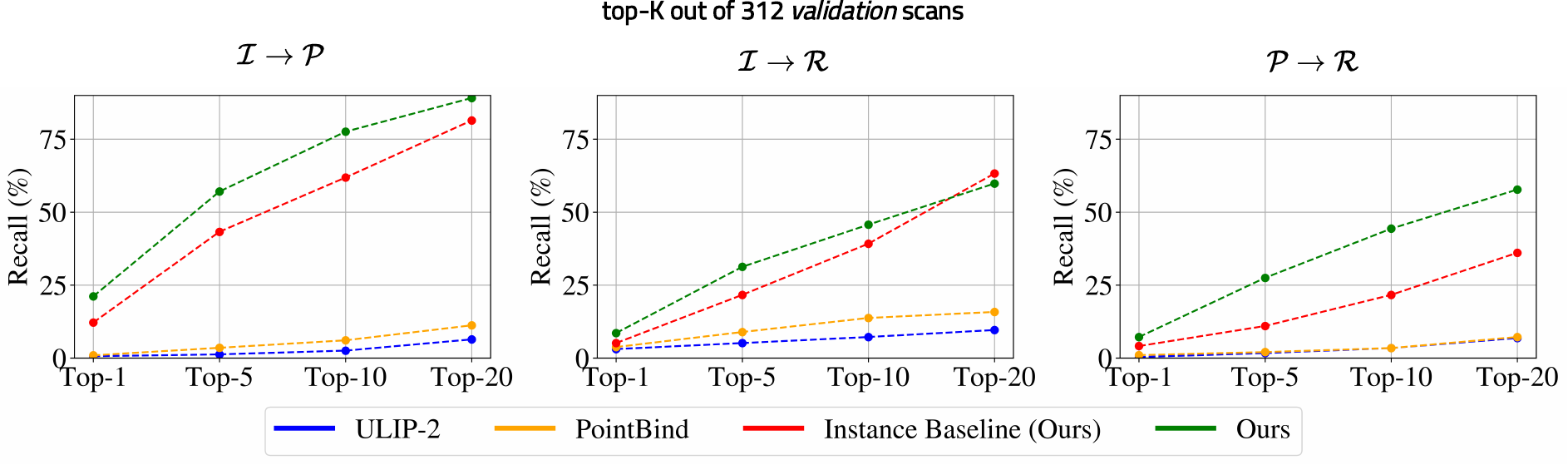}
    \vspace{-10pt}
    \caption{\textbf{Cross-Modal Scene Retrieval on \textit{ScanNet} (Scene Matching Recall).} Plots show the top 1, 5, 10, 20 scene matching recall of different methods on three modality pairs: $\mathcal{I} \rightarrow \mathcal{P}$, $\mathcal{I} \rightarrow \mathcal{R}$, $\mathcal{P} \rightarrow \mathcal{R}$. Ours and Instance Baseline have not been explicitly trained on $\mathcal{P} \rightarrow \mathcal{R}$. Results are computed on 306 scenes and showcase the superior performance of our approach. Once again, the difference between Ours and our self-baseline is attributed to the enhanced cross-modal scene-level interactions achieved with the unified encoders.}
    \label{fig:cross_modal_scene_retrieval_scannet_same_recall_graph}
\end{figure*}
\vspace{-2pt}
\begin{table}[t]
  \centering
  \resizebox{\textwidth}{!}{
   \begin{tabular}{l|ccc| ccc || ccc}
    \toprule
    \textbf{Method} & \multicolumn{3}{c}{\textbf{Scene Category Recall} $\uparrow$} & \multicolumn{3}{c}{\textbf{Temporal Recall} $\uparrow$} & \multicolumn{3}{c}{\textbf{Intra-Category Recall} $\uparrow$} \\
    \midrule\arrayrulecolor{black} 
    & \textbf{top-1} & \textbf{top-5} & \textbf{top-10} & \textbf{top-1} & \textbf{top-5} & \textbf{top-10}
    & \textbf{top-1} & \textbf{top-3} & \textbf{top-5} \\
    \midrule\arrayrulecolor{black}
    
    \multicolumn{10}{l}{\cellcolor[HTML]{EEEEEE}{\textit{$\mathcal{I} \rightarrow \mathcal{P}$}}} \\
    ULIP-2 ~\cite{xue2023ulip2} & 7.37 & 25.96 & 43.27 & 0.04 & 1.00 & 3.00 & 16.77 & 41.53 & 55.54  \\
    PointBind ~\cite{pointbind} & 13.78 & 24.36 & 42.95 & \nd 2.00 & 5.00 & 7.00 & 20.03 & 40.68 & 57.01 \\
    Inst.\ Baseline (Ours) & \nd 42.95 & \nd 70.19 & \nd 81.09 & \fs 13.00 & \nd 35.00 & \nd 60.00 & \fs 46.37 & \fs 79.68 & \fs 88.43 \\
    Ours & \fs 64.74 & \fs 89.42 & \fs 94.23 & \fs 13.00 & \fs 41.00 & \fs 84.00 & \nd 38.98 & \nd 73.28 & \nd 85.00 \\
    
    \multicolumn{10}{l}{\cellcolor[HTML]{EEEEEE}{\textit{$\mathcal{I} \rightarrow \mathcal{R}$}}} \\
    ULIP-2 ~\cite{xue2023ulip2} & 41.92 & 57.73	& 61.86	& 1.00 & 2.00 & 8.00  & 19.48 & 42.18 & 56.69 \\
    PointBind ~\cite{pointbind} & \nd 49.48	& 70.45	& \nd 80.07	& 2.00 & 6.00 & 12.00 & 19.19 & 41.54 & 55.85  \\
    Inst.\ Baseline (Ours) & 49.14 & \nd 71.48 & \nd 80.07 & \fs 8.00 & \fs 28.00 & \nd 46.00 & \nd 28.00 & \fs 62.33 & \fs 72.62 \\ 
    Ours & \fs 57.39 & \fs 82.82 & \fs 87.63 & \nd 3.00 & \nd 25.00 & \fs 51.00 & \fs 29.04 & \nd 57.85 & \nd 70.75 \\

    \multicolumn{10}{l}{\cellcolor[HTML]{EEEEEE}{\textit{$\mathcal{P} \rightarrow \mathcal{R}$}}} \\
    ULIP-2 ~\cite{xue2023ulip2} & 11.34 & 15.12 & 23.27 & \nd 1.00 & 2.00 & 4.00 & 18.12 & 41.15 & 54.93 \\
    PointBind ~\cite{pointbind} & 18.21 & 26.46 & 31.96	& \nd 1.00 & 2.00 & 6.00 & 18.25 & 40.05 & 54.84 \\
    Inst.\ Baseline (Ours) & \nd 28.87 & \nd 50.86 & \nd 66.67 & \fs 5.00 & \nd 13.00 & \nd 23.00 & \fs 29.41 & \nd 50.84 & \nd 65.65 \\ 
    Ours & \fs 57.73 & \fs 79.04 & \fs 85.57 & \fs 5.00 & \fs 20.00 & \fs 46.00 & \nd 26.79 & \fs 56.67 & \fs 68.63 \\
    \midrule
    \midrule
    
    \multicolumn{10}{l}{\cellcolor[HTML]{EEEEEE}{\textit{$\mathcal{I} \rightarrow \mathcal{F}$}}} \\
    ULIP-2 ~\cite{xue2023ulip2} & \nd 38.46 & 55.77 & 64.42 & 1.00 & 2.00 & 10.00 & 18.48 & 39.09 & 55.96  \\
    PointBind ~\cite{pointbind} & 35.58 & \nd 62.82 & \nd 72.76 & \nd 1.00 & \nd 11.00 & \nd 21.00 & \nd 20.03 & \nd 43.08 & \nd 58.62 \\
    Ours & \fs 58.01 & \fs 81.09 & \fs 89.10 & \fs 8.00 & \fs 32.00 & \fs 61.00 & \fs 28.57 & \fs 55.67 & \fs 71.77 \\

    \multicolumn{10}{l}{\cellcolor[HTML]{EEEEEE}{\textit{$\mathcal{P} \rightarrow \mathcal{F}$}}} \\ 
    ULIP-2 ~\cite{xue2023ulip2} & 13.14 & 26.28 & 33.65 & \nd 1.00 & 1.00 & 6.00 & 17.46 & 38.74 & 53.99  \\
    PointBind ~\cite{pointbind} & \nd 14.10 & \nd 48.72 & \nd 59.62	& 0.50 & \nd 5.00 & \nd 7.00 & \nd 23.17 & \nd 39.23 & \nd 57.08 \\
    Ours & \fs 55.77 & \fs 78.53 & \fs 86.54 & \fs 10.00 & \fs 30.00 & \fs 57.00 &
    \fs 31.34 & \fs 63.42 & \fs 74.15 \\
    
    \multicolumn{10}{l}{\cellcolor[HTML]{EEEEEE}{\textit{$\mathcal{R} \rightarrow \mathcal{F}$}}} \\
    ULIP-2 ~\cite{xue2023ulip2} & 8.25 & \nd 29.21 & 40.21 & \nd 1.00 & 2.00 & 5.00 & \nd 18.24 & \nd 41.80 & \nd 55.35   \\
    PointBind ~\cite{pointbind} & \nd 14.43 & 27.15 & \nd 48.45 & \nd 1.00 & \nd 5.00 & \nd 8.00 & 13.64 & 38.32 & 54.20 \\
    Ours & \fs 54.64 & \fs 74.91 & \fs 80.41 & \fs 6.00 & \fs 17.00 & \fs 35.00 & \fs 23.00 & \fs 51.37 & \fs 66.84 \\
    \bottomrule
    \end{tabular}
    }
    \caption{\textbf{Cross-Modal Scene Retrieval on \textit{ScanNet}.} We consistently outperform state-of-the-art methods and our self-baseline in most cases. The latter performs better in certain modality pairs on \textit{intra-category}, with the biggest gap observed in $\mathcal{I} \rightarrow \mathcal{R}$; this can be attributed to our less powerful text encoder.}
    \label{tab:cross_modal_scene_retrieval_scannet}
\end{table}
\begin{table}[t]
\vspace{-5pt}
  \centering
  \resizebox{\linewidth}{!}{
   \begin{tabular}{l|cccc|ccc}
    \toprule
    \textbf{Method} & \multicolumn{4}{c}{\textbf{Scene Matching Recall} $\uparrow$} & \multicolumn{3}{c}{\textbf{Temporal Recall} $\uparrow$} \\
    \midrule\arrayrulecolor{black} 
    & \textbf{top-1} & \textbf{top-5} & \textbf{top-10} & \textbf{top-20}
    & \textbf{top-1} & \textbf{top-5} & \textbf{top-10} \\
    \midrule\arrayrulecolor{black}
    
    \multicolumn{8}{l}{\cellcolor[HTML]{EEEEEE}{\textit{$\mathcal{I} \rightarrow \mathcal{P}$}}} \\
    ULIP-2 ~\cite{xue2023ulip2} & 1.27 & 5.10 & 7.01 & 12.74 & 0.04	& 4.26 & 12.77 \\ 
    PointBind ~\cite{pointbind} & 1.27 & 4.46 & 9.55 & 17.20 & \nd 2.13 & 4.26 & 8.51 \\
    Inst.\ Baseline (Ours) & \nd 8.92 & \nd 30.57 & \nd 43.31 & \nd 64.33 & 0.04 & \nd 19.15 & \nd 42.55 \\
    Ours & \fs 14.01 & \fs 49.04 & \fs 66.88 & \fs 83.44 & \fs 12.77 & \fs 36.17 & \fs 70.21 \\

    \multicolumn{8}{l}{\cellcolor[HTML]{EEEEEE}{\textit{$\mathcal{I} \rightarrow \mathcal{R}$}}} \\
    ULIP-2 ~\cite{xue2023ulip2} & 2.01 & 4.70 & 7.38 & 14.77 & \nd 2.13 & 6.38 & 12.77  \\ 
    PointBind ~\cite{pointbind} & 1.34 & 4.77 & 6.71 & 13.42 & \nd 2.13 & 6.38 & 14.89 \\
    Inst.\ Baseline (Ours) & \fs 8.72 & \fs 40.94 & \fs 57.05 & \fs 69.80 & \fs 6.38 & \fs 38.30 & \fs 63.83 \\
    Ours & \nd 6.04 & \nd 26.85 & \nd 42.28 & \nd 62.42 & \nd 2.13 & \nd 34.04 & \fs 63.83 \\

    \multicolumn{8}{l}{\cellcolor[HTML]{EEEEEE}{\textit{$\mathcal{P} \rightarrow \mathcal{R}$}}} \\
    ULIP-2 ~\cite{xue2023ulip2} & 0.67 & 3.36 & 6.71 & 12.75 & \nd 2.13 & 6.38 & 6.38 \\ 
    PointBind ~\cite{pointbind} & 0.67 & 3.36 & 6.71 & 13.42 & \nd 2.13 & 6.38 & 6.38 \\
    Inst.\ Baseline (Ours) & \nd 0.76 & \nd 14.09 & \nd 24.83 & \nd 36.24 & 0.04 & \nd 14.89 & \nd 27.66 \\
    Ours & \fs 6.71 & \fs 19.46 & \fs 32.31 & \fs 51.01 & \fs 8.51 & \fs 27.66 & \fs 51.06 \\
    \bottomrule
    \end{tabular}
    }\vspace{-5pt}
    \caption{\textbf{Cross-Modal Scene Retrieval on \textit{3RScan}.} Similar performance to the ScanNet results in Fig. \ref{fig:cross_modal_scene_retrieval_scannet_same_recall_graph} is observed.}
    \label{tab:cross_modal_scene_retrieval_scan3r}
\end{table}

\subsection{Cross-Modal Scene Retrieval}
We compare our cross-modal scene retrieval results with \cite{xue2023ulip2,pointbind} and our instance-level baseline. Since prior work does not address this task, we adapt their methods by averaging object embeddings per modality to create scene representations, treating our baseline similarly. Unlike \project{}, these methods rely on semantic instance segmentation. \textit{Scene matching recall} results on \textit{ScanNet} (Fig.~\ref{fig:cross_modal_scene_retrieval_scannet_same_recall_graph}) show that our unified encoders, not relying on semantics, consistently outperform prior methods in all pairwise modalities and surpass our baseline. Detailed results on \textit{ScanNet} and \textit{3RScan} are in Tabs.~\ref{tab:cross_modal_scene_retrieval_scannet} and~\ref{tab:cross_modal_scene_retrieval_scan3r}. Our method achieves overall scene understanding, even with small-scale object reconfigurations, as shown by its high temporal recall. The lower performance of pretrained methods may stem from training biases that limit their robustness with real-world data, such as incomplete point clouds and blurry images. Qualitative results are in Fig. \ref{fig:visual_comparison}.

\subsection{Missing Modalities}
\label{sec:missing_modalities}
To demonstrate \project{}'s ability to capture emergent modality behavior with non-overlapping training data points, we train \project{} using different data repositories for each modality pair. Specifically, we use the \textit{ScanNet} dataset and split the image repository into two chunks of varying sizes. Training on image-point cloud ($\mathcal{I} \rightarrow \mathcal{P}$) and image-mesh ($\mathcal{I} \rightarrow \mathcal{M}$) using each chunk respectively, we expect to see an emergent behavior between point cloud and mesh ($\mathcal{P} \rightarrow \mathcal{M}$). The results (Tab.~\ref{tab:ablation_miss_mod_scannet}) include top-$1$ and top-$3$ instance matching recall, as well as \textit{same} and \textit{diff} recall for evaluating intra- (e.g., identical chairs) and inter- (e.g., a chair and a table) object category performance within a scene. Although partial data availability decreases recall, our $\mathcal{P} \rightarrow \mathcal{M}$ matching only decreases by $3\%$ even when using $25\%$~$\mathcal{I} \rightarrow \mathcal{P}$. This scenario is common in real-world applications, where certain modalities might be scarce.

\begin{table}[t]
  \centering
  \resizebox{\linewidth}{!}{
   \begin{tabular}{cc|cc|cc}
    \toprule
    \multicolumn{2}{c}{\textbf{Available Data}} &  \multicolumn{4}{c}{\textbf{Instance Matching Recall} $\uparrow$} \\ 
     \midrule\arrayrulecolor{black}
      \cellcolor[HTML]{EEEEEE}{\textit{$\mathcal{I} \rightarrow \mathcal{P}$} (\%)} & \cellcolor[HTML]{EEEEEE}{\textit{$\mathcal{I} \rightarrow \mathcal{M}$} (\%) } & \textbf{same} & \textbf{diff} & \textbf{top-1} & \textbf{top-3} \\
      \midrule
      25  & 75 & 86.32 & \fs 73.38 & 55.46 & 79.73 \\
      50  & 50  & \fs 87.46 & 70.02 & \nd 57.49  & \nd 79.94 \\
      75  & 25 & 87.35 & 67.65 & 54.99 & 79.45 \\
      100 & 100 & \nd 87.44 & \nd 72.46 & \fs 59.88 & \fs 80.81 \\
    \midrule
    \end{tabular}
    }
	\caption{\textbf{Ablation on $\mathcal{P} \rightarrow \mathcal{M}$ instance matching on \textit{ScanNet} with non-overlapping data per modality pair.} Despite modality pairs not sharing the same image repository, our method retains high performance even when a pair is underrepresented in the data.}
	\label{tab:ablation_miss_mod_scannet}
 \vspace{-1mm}
\end{table}
\section{Conclusion}
\label{sec:conclusion}
In summary, this work introduces \project{}, a framework for flexible, scene-level cross-modal alignment without the need for semantic annotations or perfectly aligned data. \project{} leverages a unified embedding space centered on image features, allowing it to generalize across unpaired modalities and outperform existing methods in cross-modal scene retrieval and instance matching on real-world datasets. This approach addresses the limitations of traditional multi-modal models and holds promise for practical applications in areas like robotics, AR/VR, and construction monitoring.  Although \project{} excels in cross-modal instance matching, its scene retrieval generalizability could benefit from training on diverse indoor and outdoor datasets. \project{} assumes a base modality per dataset, advancing prior work requiring perfect modality alignment. Further relaxation is a promising direction. Finally, exploring its embedding space for downstream scene understanding remains a key area. Future research can explore how our approach can be applied to dynamic scene reconstruction and real-time navigation, thus leading to interactive and immersive mixed-reality experiences. 
\section{Acknowledgements}
\label{sec:acknowledgements}

This work was partially funded by the ETH RobotX research grant.
 
{
\small
\bibliographystyle{ieeenat_fullname}
\bibliography{main}
}

\clearpage
\setcounter{section}{0}
\counterwithin{table}{section}
\renewcommand{\thesection}{\Alph{section}}

\maketitlesupplementary

\begin{abstract}
In the supplementary material, we provide:
\begin{enumerate}[leftmargin=12pt,itemsep=0em]
    \item Impact of scaling up data (Sec.~\ref{sec:data_scaleup})
    \item Results on training with all pairwise modalities (Sec.~\ref{sec:all_pairwise_modality})
    \item Results on same modality scene retrieval (Sec.~\ref{sec:same_modal_scene_retrieval})
    \item Results on scene retrieval with one modality input to the scene-level encoder (Sec.~\ref{sec:scene_encoder_avg})
    \item Results on cross-modal coarse visual localization (Sec.~\ref{sec:sgloc_compare})
    \item Additional qualitative results on scene retrieval (Sec.~\ref{sec:qualitative_res})
    \item Details on the camera view sampling algorithm (Sec.~\ref{sec:cam_view_sampling})
    \item Analysis of inference runtime (Sec.~\ref{sec:runtime_analysis})
    \item Further details on the experimental setup (Sec.~\ref{sec:exp_details})

\end{enumerate}
\end{abstract}

\section{Data Scale-up Improvements}
\label{sec:data_scaleup}
\begin{figure*}[t]
  \begin{subfigure}[b]{0.5\linewidth}
    \centering
    \includegraphics[width=\textwidth]{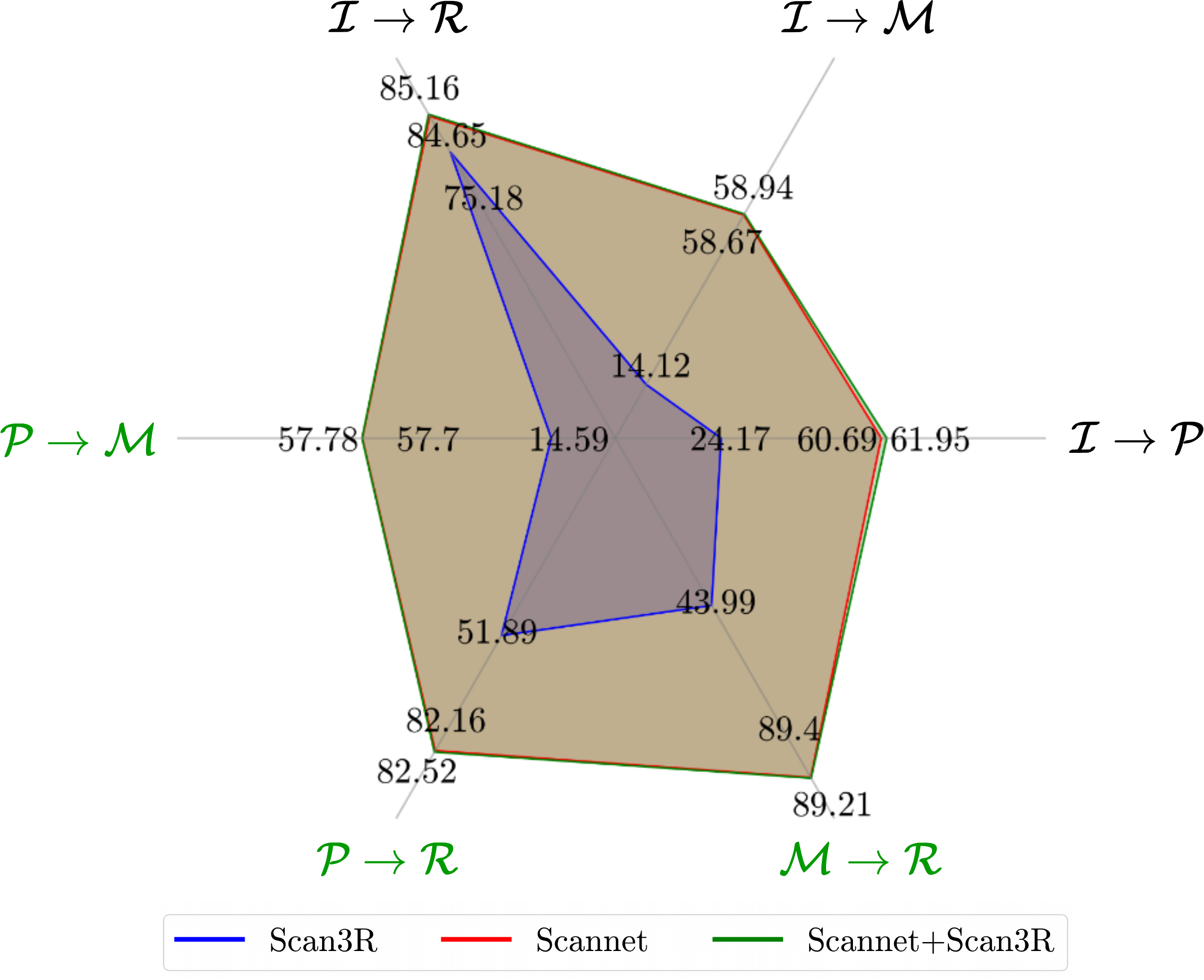}
    \caption{\textbf{Instance Matching Recall}}
    \label{fig:spider_instancematching_scaleup}
  \end{subfigure}
  \hfill
  \begin{subfigure}[b]{0.4\linewidth}
    \centering
    \resizebox{\linewidth}{!}{%
    \begin{tabular}{l|ccc}
        \toprule
        \multicolumn{4}{c}{\textbf{Scene-level Recall} $\uparrow$} \\
         \midrule\arrayrulecolor{black} 
        Trained on & \textbf{R@25}\% & \textbf{R@50}\% & \textbf{R@75}\%  \\
        \midrule
        \multicolumn{4}{l}{\cellcolor[HTML]{EEEEEE}{\textit{$\mathcal{P} \rightarrow \mathcal{M}$}}} \\
        \textit{3RScan} & 22.44 & 8.01 & 2.24 \\
        \textit{Scannet} & \fs 86.54 & \nd 64.42 & \nd 33.97 \\
        \textit{3RScan $+$ Scannet}	& \fs 86.54 & \nd 63.46 & \fs 34.29 \\

        \multicolumn{4}{l}{\cellcolor[HTML]{EEEEEE}{\textit{$\mathcal{P} \rightarrow \mathcal{R}$}}} \\
        \textit{3RScan} & 84.54 & 48.80 & 24.74 \\
        \textit{Scannet} & \fs 99.31 & \nd 96.22 & \nd 68.38 \\
        \textit{3RScan $+$ Scannet}	& \fs 99.31 & \fs 97.25 & \fs 70.10 \\

        \multicolumn{4}{l}{\cellcolor[HTML]{EEEEEE}{\textit{$\mathcal{M} \rightarrow \mathcal{R}$}}} \\
        \textit{3RScan} & 68.97 & 48.28 & 22.22 \\
        \textit{Scannet} & \fs 99.62 & \fs 98.47 & \nd 82.38 \\
        \textit{3RScan $+$ Scannet}	& \nd 99.23 & \nd 97.70 & \fs 83.91 \\
        
        \bottomrule
        \end{tabular}
    }
    \vspace{2em}
  \caption{\textbf{Scene-Level Matching Recall}}
  \label{fig:table_scene_level_instance_matching_scaleup}
  \end{subfigure}
  \caption{\textbf{Scaled-up training performance on \textit{ScanNet}.} When training on both ScanNet and 3RScan datasets together, results improve from any individual dataset training. As expected, training on 3RScan and evaluating on ScanNet will have limited performance. Note that the 3RScan includes only the $\mathcal{I}$, $\mathcal{P}$, and $\mathcal{R}$ modalities.
}
  \label{fig:instance_matching_scaleup}
\end{figure*}

We investigate the impact of scaling up training data by merging different datasets and its effect on \project{}'s performance, particularly for instance- and scene-level matching recall. Figure~\ref{fig:instance_matching_scaleup} demonstrates the advantages of joint training on the ScanNet and 3RScan datasets compared to training on each dataset individually. Please note that 3RScan includes only the $\mathcal{I}$, $\mathcal{P}$, and $\mathcal{R}$ modalities. Joint training significantly enhances scene-level recall performance and also improves instance-level recall. These results highlight CrossOver's ability to effectively leverage diverse data sources, enabling better generalization across varying scenes and object arrangements, ultimately boosting overall performance.

\section{All Pairwise Modality Training}
\label{sec:all_pairwise_modality}
As mentioned in Sec. \ref{sec:object_encoder_training} of the main paper, training with all pairwise modality combinations, as in prior work~\cite{xue2023ulip2,pointbind}, directly aligns all modality pairs in a shared embedding space. However, this approach underperforms compared to alignment with a single reference modality, as evidenced by the results in Tabs.~\ref{tab:allpairloss_comparison_scannet} and ~\ref{tab:allpairloss_comparison_scan3r}. Note that \textit{`Ours'} results are copied from Fig. \ref{fig:instance_matching} of the main paper. The key limitation of aligning all modality pairs lies in its added complexity, which dilutes focus and leads to lower scene-level recall metrics. In contrast, intra-modal alignment enhances robustness, particularly in cases of missing modality inputs, by concentrating learning on specific modality relationships. This focused alignment not only improves performance but also facilitates \textit{emergent modality} behavior. Similar insight is also noticed when training the unified encoders with the raw scene data using all pairwise modalities, namely $\mathbf{F}_{1D}$, $\mathbf{F}_{2D}$, $\mathbf{F}_{3D}$ and $\mathbf{F}_{\mathcal{S}}$. This is shown as `All Pairs' in Tabs. \ref{tab:cross_modal_scene_retrieval_scannet_scenenc} and \ref{tab:cross_modal_scene_retrieval_scan3r_scenenc}.

\begin{table}[t]
  \centering
    \begin{tabular}{l|ccc}
        \toprule
        \multicolumn{4}{c}{\textbf{Scene-level Recall} $\uparrow$} \\
         \midrule\arrayrulecolor{black} 
           & \textbf{R@25}\% & \textbf{R@50}\% & \textbf{R@75}\%  \\
        \midrule
        \multicolumn{4}{l}{\cellcolor[HTML]{EEEEEE}{\textit{$\mathcal{I} \rightarrow \mathcal{P}$}}} \\
        All Pairs & \nd 97.12 & \nd 75.00 & \nd 15.06 \\
        Ours & \fs 98.08 & \fs 76.92 & \fs 23.40 \\
        
        \multicolumn{4}{l}{\cellcolor[HTML]{EEEEEE}{\textit{$\mathcal{I} \rightarrow \mathcal{R}$}}} \\
        All Pairs & \fs 100 & \nd 98.08 & \nd 75.95 \\
        Ours & \nd 99.66 & \fs 98.28 & \fs 76.29 \\
       
        \multicolumn{4}{l}{\cellcolor[HTML]{EEEEEE}{\textit{$\mathcal{I} \rightarrow \mathcal{M}$}}} \\
        All Pairs & \fs 87.82 & \nd 63.14 & \nd 33.97 \\
        Ours & \nd 86.54 & \fs 63.46 & \fs 34.29 \\
       
        \multicolumn{4}{l}{\cellcolor[HTML]{EEEEEE}{\textit{$\mathcal{P} \rightarrow \mathcal{R}$}}} \\
        All Pairs & \fs 99.66 & \fs 97.25 & \fs 75.26 \\
        Ours \textit{\textcolor{spidergreen}{(emergent)}} & \nd 99.31 & \nd 96.56 & \nd 70.10 \\
       
        \multicolumn{4}{l}{\cellcolor[HTML]{EEEEEE}{\textit{$\mathcal{P} \rightarrow \mathcal{M}$}}} \\
        All Pairs & \fs 89.42 & \fs 65.71 & \nd 35.26 \\
        Ours \textit{\textcolor{spidergreen}{(emergent)}} & \nd 87.50 & \nd 61.54 & \nd 30.77 \\
        \multicolumn{4}{l}{\cellcolor[HTML]{EEEEEE}{\textit{$\mathcal{M} \rightarrow \mathcal{R}$}}} \\
        All Pairs & \fs 100 & \fs 98.08 & \nd 83.52 \\
        Ours \textit{\textcolor{spidergreen}{(emergent)}} & \nd 99.23 & \nd 97.70 & \fs 83.91 \\
        \bottomrule
        \end{tabular}
    \caption{\textbf{Scene-level matching results on \textit{ScanNet}.} `All Pairs' refers to training our instance-level encoder with all pairwise modality combinations. As shown, training on all pairwise combinations does not provide drastically improved performance, as one would expect, even in the modality pairs that are not directly aligned in `Ours' \textit{\textcolor{spidergreen}{(emergent)}}.}
    \label{tab:allpairloss_comparison_scannet}
\end{table}
\begin{table}[t]
  \centering
    \begin{tabular}{l|ccc}
        \toprule
        \multicolumn{4}{c}{\textbf{Scene-level Recall} $\uparrow$} \\
         \midrule\arrayrulecolor{black} 
        & \textbf{R@25}\% & \textbf{R@50}\% & \textbf{R@75}\%  \\
        \midrule
        \multicolumn{4}{l}{\cellcolor[HTML]{EEEEEE}{\textit{$\mathcal{I} \rightarrow \mathcal{P}$}}} \\
        All Pair loss & \fs 99.36 & \nd 77.71 & \nd 17.20 \\
        Ours & \fs 99.36 & \fs 79.62 & \fs 22.93 \\
        \multicolumn{4}{l}{\cellcolor[HTML]{EEEEEE}{\textit{$\mathcal{I} \rightarrow \mathcal{R}$}}} \\
        All Pair Loss & \fs 100	& \fs 97.32 & \nd 62.42 \\
        Ours & \fs 100 & \fs 97.32 & \fs 67.79 \\
        
        \multicolumn{4}{l}{\cellcolor[HTML]{EEEEEE}{\textit{$\mathcal{P} \rightarrow \mathcal{R}$}}} \\
        All Pair Loss & \fs 100 & \fs 93.96 & \fs 54.36 \\
        Ours \textit{\textcolor{spidergreen}{(emergent)}} & \fs 100 & \nd 89.26 & \nd 50.34 \\
        \bottomrule
        \end{tabular}
    \caption{\textbf{Scene-level matching results on \textit{3RScan}.} `All Pairs' refers to training our instance-level encoder with all pairwise modality combinations. Similar to ScanNet, training on all pairwise combinations does not provide improved performance, as one would expect, even in the modality pairs that are not directly aligned in `Ours' \textit{\textcolor{spidergreen}{(emergent)}}.}
    \label{tab:allpairloss_comparison_scan3r}
\end{table}

\section{Same-Modal Scene Retrieval}
\label{sec:same_modal_scene_retrieval}
We present results for \textit{same-modality scene retrieval} in Tabs.~\ref{tab:same_modal_scene_retrieval_scannet} and \ref{tab:same_modal_scene_retrieval_scan3r}, evaluated on the ScanNet and 3RScan datasets. Metrics include scene category recall, temporal recall, and intra-category recall. Our method is compared to ULIP-2~\cite{xue2023ulip2}, PointBind~\cite{pointbind}, and our instance baseline. The instance baseline is not evaluated on the floorplan modality $\mathcal{F}$ due to the lack of floorplan representation at the instance level. Additionally, the scene-level encoder combines \textit{all} instance modalities to generate the $\mathcal{F_S}$ encoding, utilizing ground truth instance segmentation that is consistent across all modalities. This can serve as an upper bound of performance for our method. 
Results indicate that individual modalities in our method are closely aligned within the embedding space, despite the cross-modal training objective. Consistent with cross-modal results, our method performs better than the instance baseline in most cases, highlighting the importance of scene-level understanding. Moreover, it achieves significantly better or comparable performance to ULIP-2 and PointBind. Notably, our method achieves 100\% accuracy on the intra-category recall metric in all modalities, consistently distinguishing the same, \eg, \textit{kitchen} among a database of \textit{kitchens}, with ULIP-2 following closely. ULIP-2 and PointBind show decreased performance on the text referral $\mathcal{R}$ modality, likely due to training on simple object descriptions (e.g., “a point cloud of a chair”) without scene context. Finally, while our scene-level encoder excels when all modalities are available, challenges arise with missing modalities, as discussed in Sec.~\ref{sec:scene_encoder_avg}.
\begin{table*}[t]
  \centering
   \begin{tabular}{l|ccc| ccc || ccc}
    \toprule
    \textbf{Method} & \multicolumn{3}{c}{\textbf{Scene Category Recall} $\uparrow$} & \multicolumn{3}{c}{\textbf{Temporal Recall} $\uparrow$} & \multicolumn{3}{c}{\textbf{Intra-Category Recall} $\uparrow$} \\
    \midrule\arrayrulecolor{black} 
    & \textbf{top-1} & \textbf{top-5} & \textbf{top-10} & \textbf{top-1} & \textbf{top-5} & \textbf{top-10}
    & \textbf{top-1} & \textbf{top-3} & \textbf{top-5} \\
    \midrule\arrayrulecolor{black}
    
    \multicolumn{10}{l}{\cellcolor[HTML]{EEEEEE}{\textit{$\mathcal{I} \rightarrow \mathcal{I}$}}} \\
    ULIP-2~\cite{xue2023ulip2} & 35.9 & 44.23 & 56.73 & 1.00 & 2.00 &  30.00 & 89.75 & 96.91 & 96.91 \\
    PointBind~\cite{pointbind} & \fs 93.59 & \nd 96.79 & \fs 98.08 & \fs 22.00 & \fs 59.00 & \fs 99.00 & \nd 90.21 & \fs 100 & \fs 100 \\
    Inst.\ Baseline (Ours)     & \nd 89.74 & 95.19 & 97.12 & \fs 22.00 & \nd 58.00 & \fs 99.00 & 80.22 & \nd 98.84 & \nd 99.87 \\
    Ours & 91.67 & \fs 97.76 & \nd 98.08 & \nd 11.00 & \fs 59.00 & \nd 98.00 & \fs 100 & \fs100 & \fs 100 \\

    \multicolumn{10}{l}{\cellcolor[HTML]{EEEEEE}{\textit{$\mathcal{R} \rightarrow \mathcal{R}$}}} \\
    ULIP-2~\cite{xue2023ulip2} & 11.34 & 18.56 & 24.05 & 1.00 & 2.00 & 4.00 & 36.63 & 57.12 & 66.17 \\
    PointBind~\cite{pointbind} & 11.34 & 18.56 & 24.05 & 1.00 & 2.00 & 4.00	& 36.63 & 57.12 & 66.17 \\
    Inst.\ Baseline (Ours) & \nd 69.42 & \fs 91.75 & \fs 94.16 & \fs 13.00 & \fs 51.00 & \fs 83.00 & \nd 86.56 & \nd 97.65 & \nd 99.20 \\
    Ours & \fs 76.98 & \nd 91.75 & \nd 94.85 & \fs 14.00 & \nd 40.00 & \nd 79.00 & \fs 100 & \fs 100 & \fs 100 \\
    \multicolumn{10}{l}{\cellcolor[HTML]{EEEEEE}{\textit{$\mathcal{P} \rightarrow \mathcal{P}$}}} \\
    ULIP-2~\cite{xue2023ulip2} & 13.14 & 13.14 & 23.72 & 1.00 & 2.00 & 3.00 & 21.52 & 42.12 & 57.25  \\
    PointBind~\cite{pointbind} & 17.63 & 58.33 & 71.47 & 7.00 & 23.00 & 45.00 & 59.54 & 90.36 & 96.46 \\
    Inst.\ Baseline (Ours) & \nd 38.14 & \nd 75.00 & \nd 85.38 & \nd 14.00 & \nd 42.00 & \nd 73.00 & \nd 86.31 & \nd 97.14 & \nd 99.81 \\
    Ours & \fs 86.54 & \fs 95.51 & \fs 96.79 & \fs 19.00 & \fs 57.00 & \fs 96.00 & \fs 100 & \fs 100 & \fs 100 \\
    \midrule
    \midrule
    \multicolumn{10}{l}{\cellcolor[HTML]{EEEEEE}{\textit{$\mathcal{F} \rightarrow \mathcal{F}$}}} \\
    ULIP-2~\cite{xue2023ulip2} & 13.78 & 24.36 & 41.03 & 1.00 & 2.00 & 5.00 & \nd 99.27 & \nd 99.89 & \nd 99.89 \\
    PointBind~\cite{pointbind} & \nd 63.78 & \nd 82.37 & \nd 89.10 & \nd 7.00 & \nd 37.00 & \nd 67.00 & \fs 100 & \fs 100 & \fs 100 \\
    Ours & \nd 59.95 & \fs 83.65 & \fs 90.38 & \fs 14.00 & \fs 43.00 & \fs 74.00 & \fs 100 & \fs 100 & \fs 100 \\
    \midrule
    \midrule
    \multicolumn{10}{l}{\cellcolor[HTML]{EEEEEE}{\textit{$\mathbf{F}_{\mathcal{S}} \rightarrow \mathbf{F}_{\mathcal{S}}$}}} \\
    Ours & 94.23 & 97.44 & 98.08 & 17.00 & 57.00 & 99.00 & 100 & 100 & 100 \\
    \bottomrule
    \end{tabular}
    \caption{\textbf{Same-Modality Scene Retrieval on \textit{ScanNet}.} Our method performs on par with or better than baselines in same-modality scene retrieval across most metrics, indicating that individual modalities in our method are closely aligned within the embedding space, despite the cross-modal training objective.}
    \label{tab:same_modal_scene_retrieval_scannet}
\end{table*}
\begin{table}[t]
  \centering
   \begin{tabular}{l|ccc}
    \toprule
    \textbf{Method} & \multicolumn{3}{c}{\textbf{Temporal Recall} $\uparrow$} \\
    \midrule\arrayrulecolor{black} 
    & \textbf{top-1} & \textbf{top-5} & \textbf{top-10} \\    \midrule\arrayrulecolor{black}
    \multicolumn{4}{l}{\cellcolor[HTML]{EEEEEE}{\textit{$\mathcal{I} \rightarrow \mathcal{I}$}}} \\
    ULIP-2~\cite{xue2023ulip2} & 2.13 & 8.51 & 29.79 \\
    PointBind~\cite{pointbind} & \nd 10.64 & 51.06 & 93.62 \\
    Inst.\ Baseline (Ours) & 4.26 & \fs 65.96 & \fs 100 \\
    Ours & \fs 17.02 & \nd 61.70 & \fs 100 \\
    
    \multicolumn{4}{l}{\cellcolor[HTML]{EEEEEE}{\textit{$\mathcal{R} \rightarrow \mathcal{R}$}}} \\
    ULIP-2~\cite{xue2023ulip2} & 2.13 & 6.38 & 8.51 \\
    PointBind~\cite{pointbind} & 2.13 & 6.38 & 8.51 \\
    Inst.\ Baseline (Ours) & \fs 19.15 & \fs 46.81 & \nd 91.49 \\
    Ours & \nd 12.77 & \nd 51.06 & \fs 87.23 \\
    
    \multicolumn{4}{l}{\cellcolor[HTML]{EEEEEE}{\textit{$\mathcal{P} \rightarrow \mathcal{P}$}}} \\
    ULIP-2~\cite{xue2023ulip2} & 0.04 & 4.26 & 6.38 \\
    PointBind~\cite{pointbind} & 2.13 & 17.02 & 36.17 \\
    Inst.\ Baseline (Ours) & \nd 6.38 & \nd 29.79 & \nd 3.83 \\
    Ours & \fs 19.15 & \fs 65.96 & \fs 97.87 \\    
    \midrule
    \midrule
    \multicolumn{4}{l}{\cellcolor[HTML]{EEEEEE}{\textit{$\mathbf{F}_{\mathcal{S}} \rightarrow \mathbf{F}_{\mathcal{S}}$}}} \\
    Ours & 17.02 & 59.57 & 97.87 \\
    \bottomrule
    \end{tabular}
    \caption{\textbf{Same-Modality Scene Retrieval on \textit{3RScan}.} Our method performs on par with or better than baselines in same-modality scene retrieval across most metrics. The lower performance on this dataset is likely due to limited training data availability.}
    \label{tab:same_modal_scene_retrieval_scan3r}
\end{table}

\begin{figure*}
    \centering
    \includegraphics[width=\columnwidth]{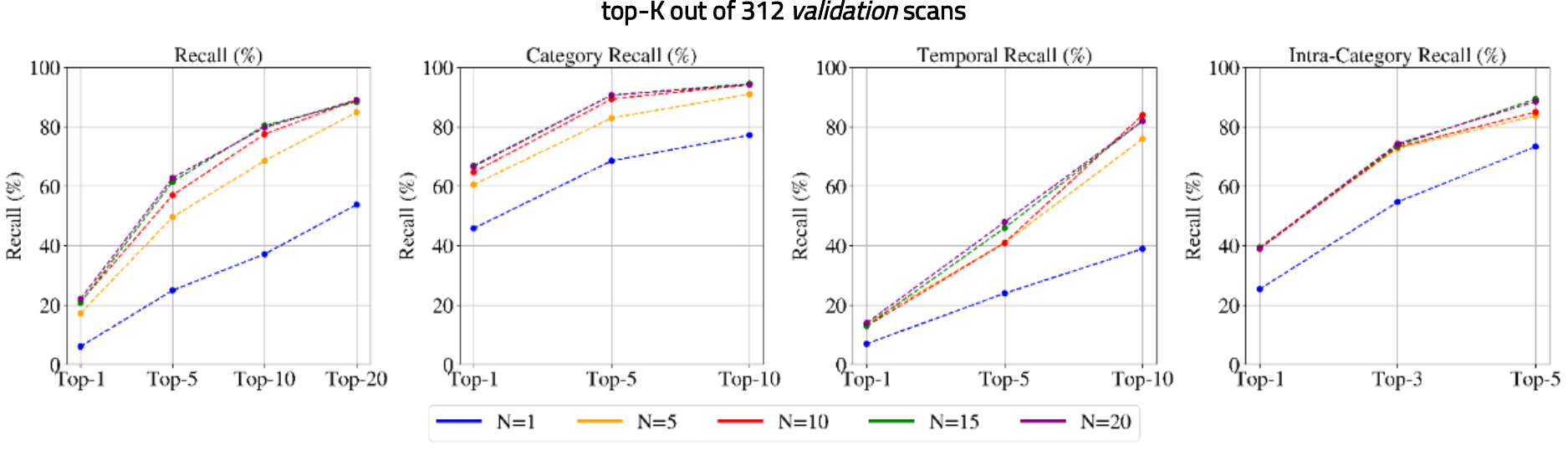}
    \vspace{-20pt}
    \caption{\textbf{Cross-Modal $\mathcal{I} \rightarrow \mathcal{P}$ Scene Retrieval on \textit{ScanNet}.} Plots showcase scene matching recall (Recall), category recall, temporal recall, and intra-category recall for different number of camera views evaluated on several Top-$k$ matches. Note that maximum $k$ differs per recall since the amount of eligible matches depends on the criteria for each recall type: scene similarity, category, temporal changes.}
    \label{fig:camera_view_abl_scannet}
\end{figure*}

\section{Uni-modal Scene-Level Encoder Inference}
\label{sec:scene_encoder_avg}
\begin{table*}
  \centering
  \resizebox{\textwidth}{!}{
   \begin{tabular}{l| cccc | ccc| ccc || ccc}
    \toprule
    \textbf{Method} & \multicolumn{4}{c}{\textbf{Scene Matching Recall}  $\uparrow$} & \multicolumn{3}{c}{\textbf{Scene Category Recall} $\uparrow$} & \multicolumn{3}{c}{\textbf{Temporal Recall} $\uparrow$} & \multicolumn{3}{c}{\textbf{Intra-Category Recall} $\uparrow$} \\
    \midrule\arrayrulecolor{black} 
    & \textbf{top-1} & \textbf{top-5} & \textbf{top-10} & \textbf{top-20} & \textbf{top-1} & \textbf{top-5} & \textbf{top-10} &
    \textbf{top-1} & \textbf{top-5} & \textbf{top-10} &
    \textbf{top-1} & \textbf{top-3} & \textbf{top-5} \\
    \midrule\arrayrulecolor{black}
    \multicolumn{14}{l}{\cellcolor[HTML]{EEEEEE}{\textit{$\mathcal{I} \rightarrow \mathcal{P}$}}} \\
    Uni-modal & \nd 16.67 & \nd 51.92 & 66.67 &  85.26 & 36.22 &  73.72 &  85.26 & \fs 14.00 & 36.00 & 67.00 & \fs 49.05 & \fs 85.15 & \fs 91.91 \\
    All Pairs & 16.35 & 54.17 & \nd 75.32 & \fs 91.35 & \fs 65.71 & \nd 86.54 & \nd 93.91 & 11.00 & \fs 42.00 & \nd 77.00 & 41.51 & 71.38 & 84.85 \\
    Ours & \fs 21.15 & \fs 57.05 & \fs 77.56 & \nd 89.10 & \nd 64.74 & \fs 89.42 & \fs 94.23 & \nd 13.00 & \nd 41.00 & \fs 84.00 & \nd 38.98 & \nd 73.28 & \nd 85.00 \\

     \multicolumn{14}{l}{\cellcolor[HTML]{EEEEEE}{\textit{$\mathcal{I} \rightarrow \mathcal{R}$}}} \\
     Uni-modal & 2.75 & 11.00 & 18.21 & 29.90 & 19.59 & 46.74 & 62.89 & 2.00 & 14.00 & 19.00 & 26.12 & 55.80 & 66.71 \\
     All Pairs & \nd 7.56 & \fs 33.68 & \fs 50.17 & \fs 65.64 & \fs 65.98 & \fs 83.16 & \fs 88.66 & \fs 8.00 & \fs 28.00 & \fs 52.00 & \fs 29.99 & \fs 58.42 & \fs 72.64 \\
    Ours & \fs 8.59 & \nd 31.27 & \nd 45.70 & \nd 59.79 & \nd 57.39 & \nd 82.82 & \nd 87.63 & \nd 3.00 & \nd 25.00 & \nd 51.00 & \nd 29.04 & \nd 57.85 & \nd 70.75 \\

     \multicolumn{14}{l}{\cellcolor[HTML]{EEEEEE}{\textit{$\mathcal{P} \rightarrow \mathcal{R}$}}} \\
    Uni-modal & 2.06 & 5.15 & 12.03 & 21.31 & 11.68	& 39.86	& 57.04	& 3.00 & 6.00 & 11.00 & 25.82 & 53.52	& 68.08 \\
    All Pairs & \nd 6.87 & \nd 24.05 & \nd 37.46 & \fs 58.42 & \nd 56.70 & \nd 74.57 & \nd 82.82 & \nd 3.00	&  \fs 22.00 & \nd 41.00 & \fs 31.94 & 	\nd 56.12 & \fs 70.22 \\
    Ours & \fs 7.22 & \fs 27.49 & \fs 44.33 & \nd 57.73 & \fs 57.73 & \fs 79.04 & \fs 85.57 & \fs 5.00 & \nd 20.00 & \fs 46.00 & \nd 26.79 & \fs 56.57 &  \nd 68.63 \\
     
    \bottomrule
    \end{tabular}
    }
    \caption{\textbf{Uni-modal \& All pair-wise modality training on Scene-Level Encoder Inference on Cross-Modal Scene Retrieval on \textit{ScanNet}.} `All Pairs' refers to training our unified encoder with all pairwise modality combinations. `Uni-modal' refers to the scene-level encoder with single-modality input. As shown in the Table, our approach outperforms the scene-level encoder and `All Pairs' in most cases. Unlike the unified dimensionality encoders, the scene-level encoder relies on instance-level data, even when operating on a single modality.}
    \label{tab:cross_modal_scene_retrieval_scannet_scenenc}
\end{table*}
In Sec.~\ref{sec:unified_encoder_training} of the main paper, we highlighted two key advantages of unified dimensionality encoders over the scene-level encoder: (i) they eliminate the need for instance-level modalities or instance information, and (ii) the scene-level encoder struggles when provided with only a single modality (uni-modal) instead of all. To validate the latter, cross-modal scene retrieval results are presented in Tabs. \ref{tab:cross_modal_scene_retrieval_scannet_scenenc} and \ref{tab:cross_modal_scene_retrieval_scan3r_scenenc}. Our method significantly outperforms the uni-modal scene-level encoder in most cases, underscoring the effectiveness and value of the unified modality encoders.
\begin{table*}[t]
  \centering
  \resizebox{0.6\textwidth}{!}{
   \begin{tabular}{l|cccc|ccc}
    \toprule
    \textbf{Method} & \multicolumn{4}{c}{\textbf{Scene Matching Recall}  $\uparrow$} & \multicolumn{3}{c}{\textbf{Temporal Recall} $\uparrow$} \\
    \midrule\arrayrulecolor{black} 
    & \textbf{top-1} & \textbf{top-5} & \textbf{top-10} & \textbf{top-20} & \textbf{top-1} & \textbf{top-5} & \textbf{top-10} \\
    \midrule\arrayrulecolor{black}
    \multicolumn{8}{l}{\cellcolor[HTML]{EEEEEE}{\textit{$\mathcal{I} \rightarrow \mathcal{P}$}}} \\
    Uni-modal & 11.46 & 42.68 & 64.33 & 84.71 & 12.77 & 31.91 & 68.09 \\
    All Pairs & \nd 12.74 & \nd 43.31 & \nd 64.97 & \nd 80.89 & \nd 8.51 & \fs 44.68 & \fs 74.47 \\
    Ours &  \fs 14.01 & \fs 49.04 & \fs 66.88 & \fs 83.44 & \fs 12.77 & \nd 36.17 & \nd 70.21\\
     \multicolumn{8}{l}{\cellcolor[HTML]{EEEEEE}{\textit{$\mathcal{I} \rightarrow \mathcal{R}$}}} \\
     Uni-modal & 3.36 & 14.77 & 28.86 & 51.01 & 8.51 & 21.28 & 42.55 \\
     All Pairs & \fs 8.05 & \fs 30.20 & \fs 46.98 & \nd 60.40 & \fs 8.51 & \nd 31.91 & \nd 59.57 \\
      Ours &   \nd 6.04	& \nd 26.85 & \nd 42.28 & \fs 62.42 & \nd 2.13 & \fs 34.04 & \fs 63.83 \\

     \multicolumn{8}{l}{\cellcolor[HTML]{EEEEEE}{\textit{$\mathcal{P} \rightarrow \mathcal{R}$}}} \\
     Uni-modal & 1.34 & 12.08 & 19.46 & 36.91 & 4.26 & 14.89 & 29.79 \\
     All Pairs & \fs 7.38 & \fs 21.48 & \fs 37.58 & \fs 59.73 & \nd 4.26 & \fs 29.79 & \fs 55.32 \\
     Ours & \nd 6.71 & \nd 19.46 & \nd 32.21 & \nd 51.01 & \fs 8.51 & \nd 27.66 & \nd 51.06 \\
     \bottomrule
    \end{tabular}
    }
    \caption{\textbf{Uni-modal \& All pair-wise modality training on Scene-Level Encoder Inference on Cross-Modal Scene Retrieval on \textit{3RScan}.} `All Pairs' refers to training our unified encoder with all pairwise modality combinations. `Uni-modal' refers to the scene-level encoder with single-modality input. As shown in the Table, our approach outperforms the scene-level encoder in all but one case. Unlike the unified dimensionality encoders, the scene-level encoder relies on instance-level data, even when operating with a single modality.}
    \label{tab:cross_modal_scene_retrieval_scan3r_scenenc}
\end{table*}

\section{Cross-Modal Coarse Visual Localization}
\label{sec:sgloc_compare}
\begin{table*}[t]
  \centering
  \resizebox{0.6\textwidth}{!}{
   \begin{tabular}{l| ccc | ccc}
    \toprule
    \textbf{Method} & \multicolumn{6}{c}{\textbf{Static Scenario}} \\
    \midrule\arrayrulecolor{black} 
    & \multicolumn{3}{c}{\textbf{R out of $10$}  $\uparrow$}
    & \multicolumn{3}{c}{\textbf{R out of $50$}  $\uparrow$} \\   
    \midrule\arrayrulecolor{black} 
    & \textbf{top-1} & \textbf{top-5} & \textbf{top-10} &
    \textbf{top-1} & \textbf{top-5} & \textbf{top-10} \\
    \midrule\arrayrulecolor{black} 
    LidarCLIP~\cite{lidarclip2024} & 16.30 & 41.40 & 60.60 & 4.70 & 11.00 & 16.30 \\
    LipLoc~\cite{shubodh2024lip} & 14.00 & 35.80 & 57.90 & 2.00 & 8.60 & 15.20 \\
    SceneGraphLoc~\cite{miao2024scenegraphloc} & \fs53.60 & \fs81.90 & \fs92.80 & \fs30.20 & \fs50.20 & \fs61.20 \\
    Ours & \nd46.00 & \nd77.97 & \nd90.58 & \nd18.69 & \nd39.16 & \nd51.62 \\
    \bottomrule
    \end{tabular}
    }
    \caption{\textbf{Cross-Modal Coarse Visual Localization on \textit{3RScan}.} Given a single image of a scene, the goal is to retrieve the corresponding scene from a database of multi-modal maps. We evaluate following the experimental setup in SceneGraphLoc~\cite{miao2024scenegraphloc} and compare our method to it and its baselines. Despite encoding less information in our multi-modal maps, our method performs competitively with SceneGraphLoc.}
    \label{tab:sgloc_comparison}
\end{table*}

We evaluate our method on the task of cross-modal coarse visual localization of a single image against a database of multi-modal reference maps, comparing it to SceneGraphLoc~\cite{miao2024scenegraphloc} and its baselines LipLoc~\cite{shubodh2024lip} and LidarCLIP~\cite{lidarclip2024} on the 3RScan dataset. SceneGraphLoc uses 3D scene graphs during inference as the multi-modal reference maps, incorporating object instance point clouds, their attributes and relationships, and the scene's structure (for a formal definition of these modalities we point the reader to~\cite{sarkar2023sgaligner,miao2024scenegraphloc}). For a fair comparison, we use the 2D unified dimensionality encoder to process the input image into an $\mathcal{F}_{2D}$ feature vector, which is then compared to the $\mathcal{F}_{S}$ feature vectors of all scenes in the database, extracted by our scene-level encoder. As shown in Tab. \ref{tab:sgloc_comparison}, despite encoding less information in our multi-modal maps, our method performs competitively with SceneGraphLoc.

\section{Qualitative Results}
\label{sec:qualitative_res}

\begin{figure*}
    \centering
    \includegraphics[width=\linewidth]{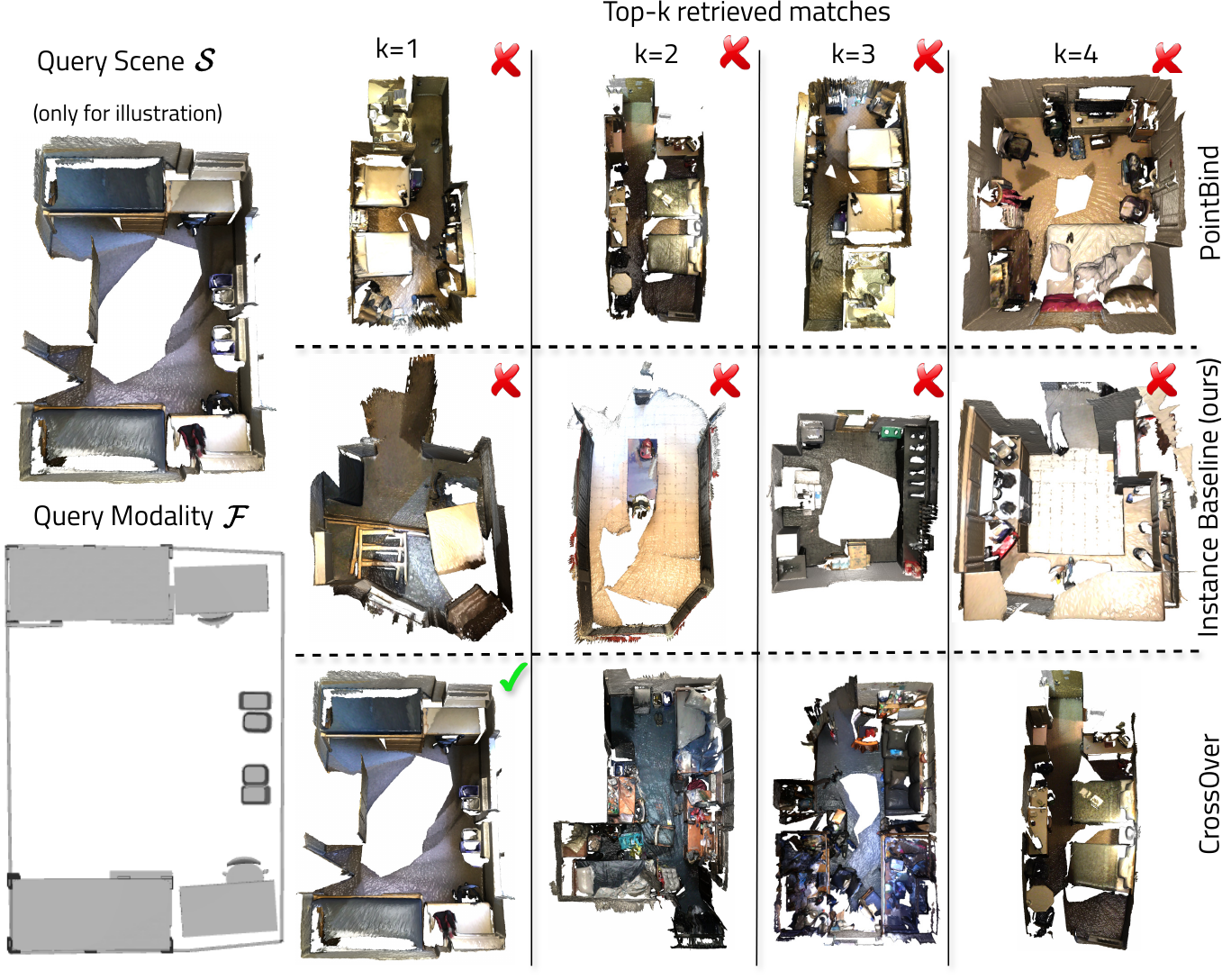}
    \caption{\textbf{Cross-Modal Scene Retrieval \textcolor{spidergreen}{Success} Qualitative Results on ScanNet.} Given a scene in query modality $\mathcal{F}$, we aim to retrieve the same scene in target modality $\mathcal{P}$. While PointBind and the Instance Baseline do not retrieve the correct scene within the top-4 matches, \project{} identifies it as the top-1 match.}
    \label{fig:visual_comparison_add_success}
    \vspace{-5pt}
\end{figure*}

\begin{figure*}
    \centering
    \includegraphics[width=\linewidth]{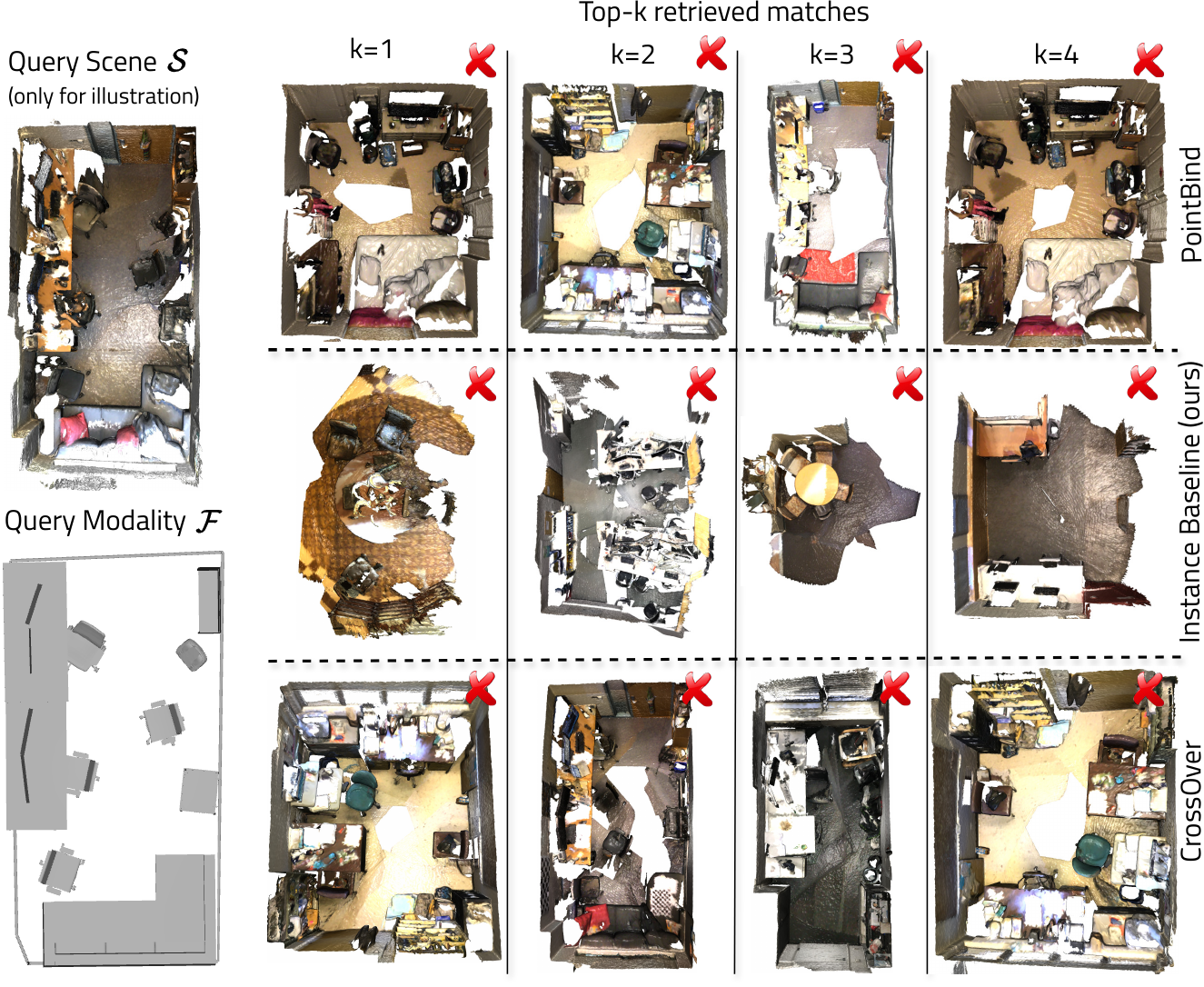}
    \caption{\textbf{Cross-Modal Scene Retrieval \textcolor{red}{Failure} Qualitative Results on ScanNet.} Given a scene in query modality $\mathcal{F}$, we aim to retrieve the same scene in target modality $\mathcal{P}$. While the baselines also fail to retrieve the same scene, CrossOver ($k=2$) and PointBind ($k=3$) retrieve a temporal scan as match.}
    \label{fig:visual_comparison_add_failure}
    \vspace{-5pt}
\end{figure*}

\begin{figure*}
    \centering
    \includegraphics[width=\linewidth]{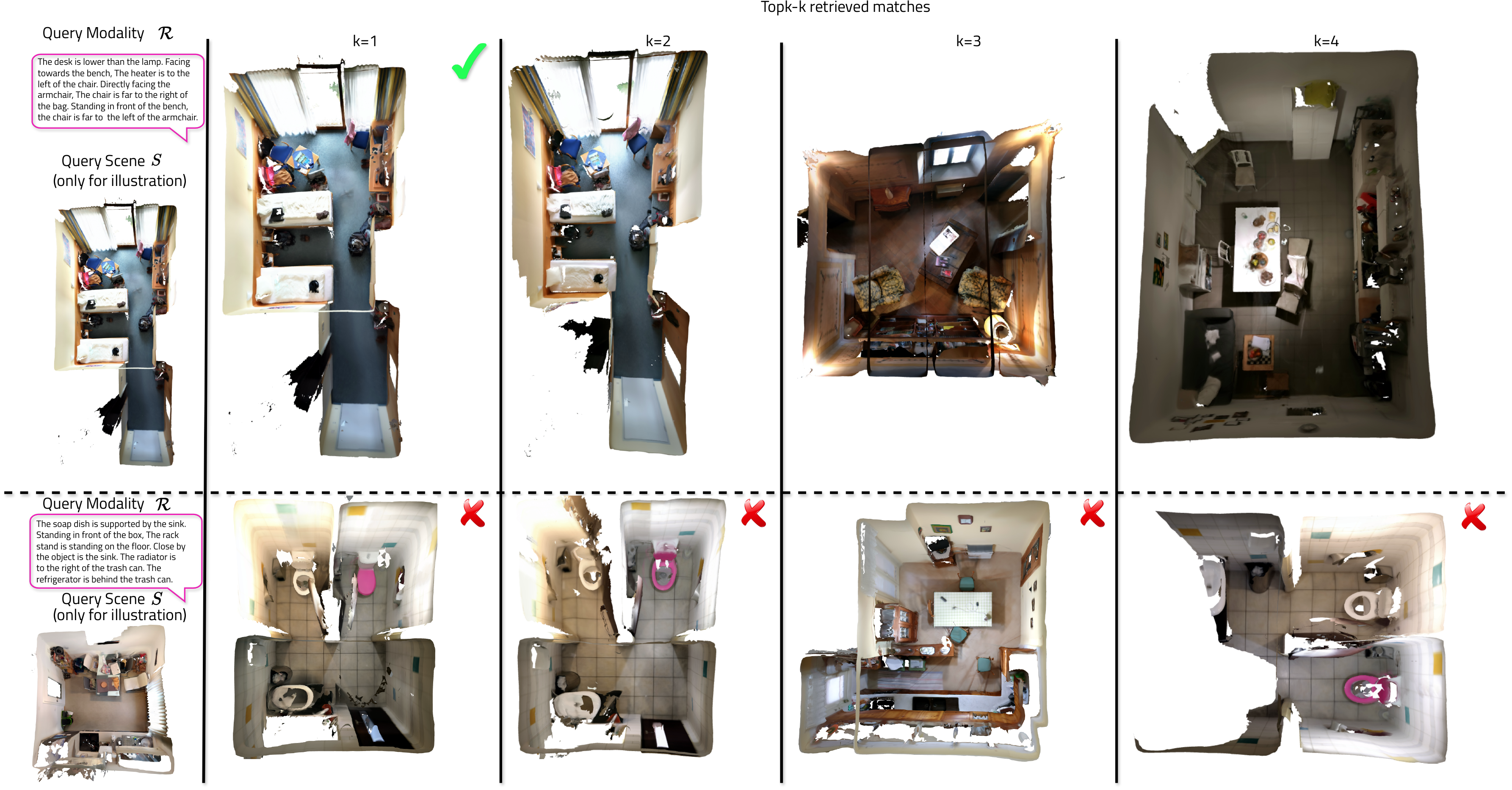}
    \caption{\textbf{Cross-Modal Scene Retrieval Qualitative Results on 3RScan. Top row - \textcolor{spidergreen}{Success}, Bottom row - \textcolor{red}{Failure}.} Given a scene in query modality $\mathcal{R}$, we aim to retrieve the same scene in target modality $\mathcal{P}$. Temporal scenes cluster naturally in the embedding space. However, query referrals may retrieve scans with similar objects across different scenes, especially when not discriminative enough (bottom).}
    \label{fig:visual_comparison_scan3r}
    \vspace{-5pt}
\end{figure*}

We present additional qualitative results in Figs. \ref{fig:visual_comparison_add_success} and \ref{fig:visual_comparison_add_failure} for cross-modal scene retrieval of the pairwise modalities $\mathcal{F} \rightarrow \mathcal{P}$. Fig. \ref{fig:visual_comparison_add_success} illustrates a success case for our method, where the correct scene is retrieved in the first match. In contrast, PointBind~\cite{pointbind} and our instance baseline fail to retrieve the correct scene within the first four matches. Notably, our instance baseline does not retrieve any bedrooms. Fig. \ref{fig:visual_comparison_add_failure} illustrates a failure case of our method. Despite this, it successfully retrieves all office scenes with a layout similar to the query one. In comparison, the baselines also fail to retrieve the correct scene but instead find matches in bedrooms and meeting rooms. Fig. \ref{fig:visual_comparison_scan3r} shows success and failure cases on 3RScan dataset for cross-modal scene retrieval of the pairwise modalities $\mathcal{R} \rightarrow \mathcal{P}$.

\begin{figure}
    \centering
    \includegraphics[width=\columnwidth]{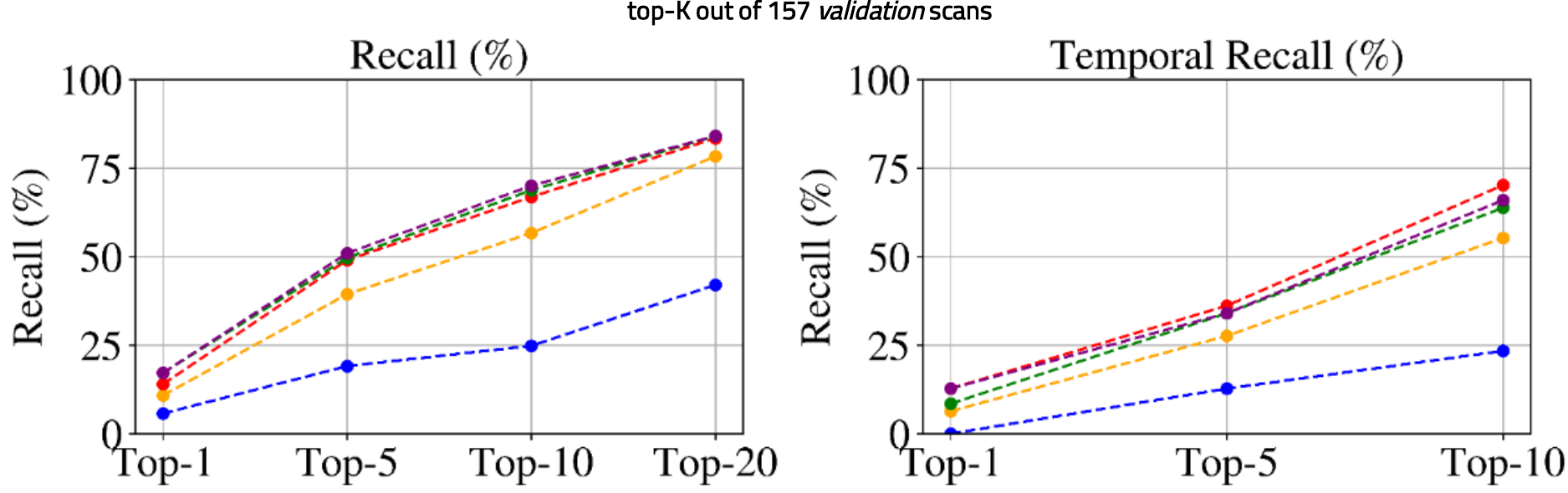}
    \vspace{-20pt}
    \caption{\textbf{Cross-Modal $\mathcal{I} \rightarrow \mathcal{P}$ Scene Retrieval on \textit{3RScan}.} Plots showcase scene matching recall (Recall) and temporal recall for different number of camera views.}
    \label{fig:camera_view_abl_scan3r}
\end{figure}
\vspace{-2pt}

\section{Camera View Sampling}
\label{sec:cam_view_sampling}

To sample camera views for the unified 2D encoder (Sec. \ref{sec:unified_encoder_training} of the main paper), we represent each camera pose as a $7D$ grid, combining its $3D$ translation and quaternion-based rotation ($4$ quaternion $+$ $3$ translation components). Our method selects $N$ camera poses to maximize $3D$ spatial separation in rotation and translation. Starting with a random pose, we iteratively select the pose farthest from \textit{all} previously chosen ones. This method ensures an even and diverse sampling of camera viewpoints across the scene. We analyze the impact of the number of selected cameras and present results for $N$ values of $1, 5, 10$, and $20$) in Figs. \ref{fig:camera_view_abl_scannet} and \ref{fig:camera_view_abl_scan3r}. The results show that performance stabilizes after $N=10$, with additional frames providing only slight improvements, indicating full scene coverage is not necessary for training \project{}. Consequently, we use $N=10$ for all reported results in our method.

\begin{figure*}
    \centering
    \includegraphics[width=\linewidth]{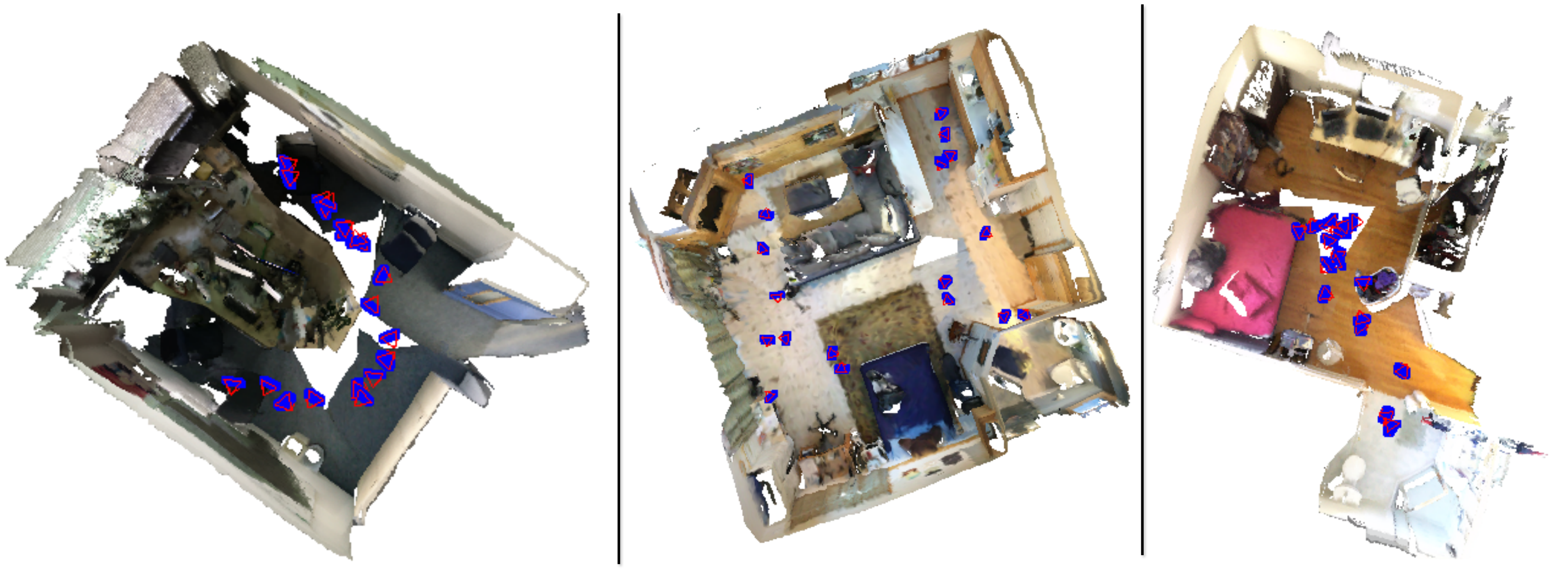}
    \caption{\textbf{Camera View Sampling Visualisation on ScaNnet dataset.} Here, we visualise the $N=20$ selected views (in \textcolor{purple}{purple} projected onto the ground truth scene mesh) using  our camera sampling method. Although, the selected cameras may not cover the entire scene, they are spatially well-separated.}
    \label{fig:camera_view_sampling}
\end{figure*}  

\section{Runtime Analysis}
\label{sec:runtime_analysis}
Our scene retrieval model consists of $1.5$B-parameter. On an NVIDIA 4090 GPU, our model takes $1.01$s $\pm 0.26$s for a single modality and $1.98$s for all modalities in $1D$, $2D$ and $3D$. It can be adapted to lightweight encoders for faster inference in compute-limited scenarios, with potential performance trade-off.

\section{Experimental Setup Details}
\label{sec:exp_details}
\noindent \textbf{Location Encoding \& Instance Spatial Relationships.} Given $\mathcal{P}_i$, we compose features $f_i^\mathcal{P}$ and the location $l_i$ (ie, $3$D position, length, width and height) to form instance tokens $\hat{f_i^\mathcal{P}}$~\cite{3dvista}. A similar process is followed for $\mathcal{M}_i$. Since we do not use scene graph representations, for instance modality $\mathcal{P}$, we embed the pairwise spatial relationships between objects in a spatial transformer~\cite{jia2024sceneverse,3dvista} to encode the scene layout and context. For any two objects $\mathcal{O}_i$ and $\mathcal{O}_j$ present in a scene, we define relationship $s_{ij} = [d_{ij}, sin(\theta_h), cos(\theta_h), sin(\theta_v), cos(\theta_v)]$, where $d_{ij}$ is the Euclidean distance between the centroids of objects $i$ and $j$, and $\theta_h$ and $\theta_v$ are the horizontal and vertical angles of the line connecting the centers of objects $i$ and $j$. The pairwise spatial feature matrix $S = \{s_{ij}\}$ is used to update the attention weights in the self-attention layers of the transformer.

\noindent \textbf{Evaluation Setup.} Our results are reported on the \textit{validation} sets of ScanNet \cite{dai2017scannet} and 3RScan \cite{wald2019rio}, as their corresponding \textit{test} sets lack public annotations or is unavailable. For the experiments in Sec.~\ref{sec:sgloc_compare}, we follow the dataset split provided by SceneGraphLoc~\cite{miao2024scenegraphloc} to ensure fairness.

\noindent \textbf{Implementation.} Inspired by CLIP~\cite{Radford2021LearningTV}, we adopt an embedding space of size $768$, consistent across instance-level, scene-level, and unified training stages. Each model is trained for 300 epochs on an NVIDIA GeForce RTX 4090 Ti GPU using the AdamW optimizer~\cite{Loshchilov2017DecoupledWD} with a learning rate of $1e-3$, and a cosine annealing scheduler with warm restarts. To fine-tune the pre-trained encoders (BLIP~\cite{Li2022BLIPBL}, DinoV2~\cite{oquab2023dinov2, darcet2023vitneedreg}, and I2PMAE \cite{Zhang2022Learning3R}), we employ a 2-layer MLP projection head with dropout and Layer Normalization~\cite{miao2024scenegraphloc, girdhar2023imagebind}. The $\tau$ parameter in the contrastive loss formulation is treated as a learnable parameter. Consistent with practices in \cite{jia2024sceneverse}, we pre-train object-level and scene-level encoders and freeze them during unified dimensionality encoder training. 

\end{document}